\documentclass{article}

\usepackage{microtype}
\usepackage{graphicx}
\usepackage{subfigure}
\usepackage{booktabs} % for professional tables

\usepackage{hyperref}

\usepackage[accepted]{icml2018}

\usepackage{adjustbox}
\usepackage{array}
\usepackage{multirow}

\newcommand{\COMMENT}[1]{}
\usepackage{amsmath,amsthm}

\begin{document}
% \pagenumbering{arabic}

\twocolumn[
%\icmltitle{Learning Class-Topology based on the Wallpaper Group Hierarchy}
\icmltitlerunning{EscherNet 101}
\icmltitle{\LARGE EscherNet 101}



\icmlsetsymbol{equal}{*}

\begin{icmlauthorlist}
\icmlauthor{Christopher Funk}{kitware}
\icmlauthor{Yanxi Liu}{psu}
\end{icmlauthorlist}

\icmlaffiliation{kitware}{Kitware (work done while at PSU)}
\icmlaffiliation{psu}{School of Electrical Engineering and Computer Science}
% \icmlaffiliation{goo}{Googol ShallowMind, New London, Michigan, USA}
% \icmlaffiliation{ed}{School of Computation, University of Edenborrow, Edenborrow, United Kingdom}
%
\icmlcorrespondingauthor{Christopher Funk}{christopher.funk@kitware.com}
\icmlcorrespondingauthor{Yanxi Liu}{yul11@psu.edu}

%\icmlkeywords{Machine Learning, ICML}

\vskip 0.3in
]



\printAffiliationsAndNotice{}  % leave blank if no need to mention equal contribution

\begin{abstract} 
	A deep learning model, EscherNet 101, is constructed to categorize images of 2D periodic patterns into their respective 17 wallpaper groups. 
    Beyond evaluating EscherNet 101 performance by classification rates,  
    at a micro-level we investigate the filters learned at different layers in the network, capable of capturing second-order invariants beyond edge and curvature.
\end{abstract}

\section{Introduction} 
\label{sec:Introduction}

\begin{figure}[!b]
	\centering 
	\vspace{-10pt}
\includegraphics[width=1\linewidth]{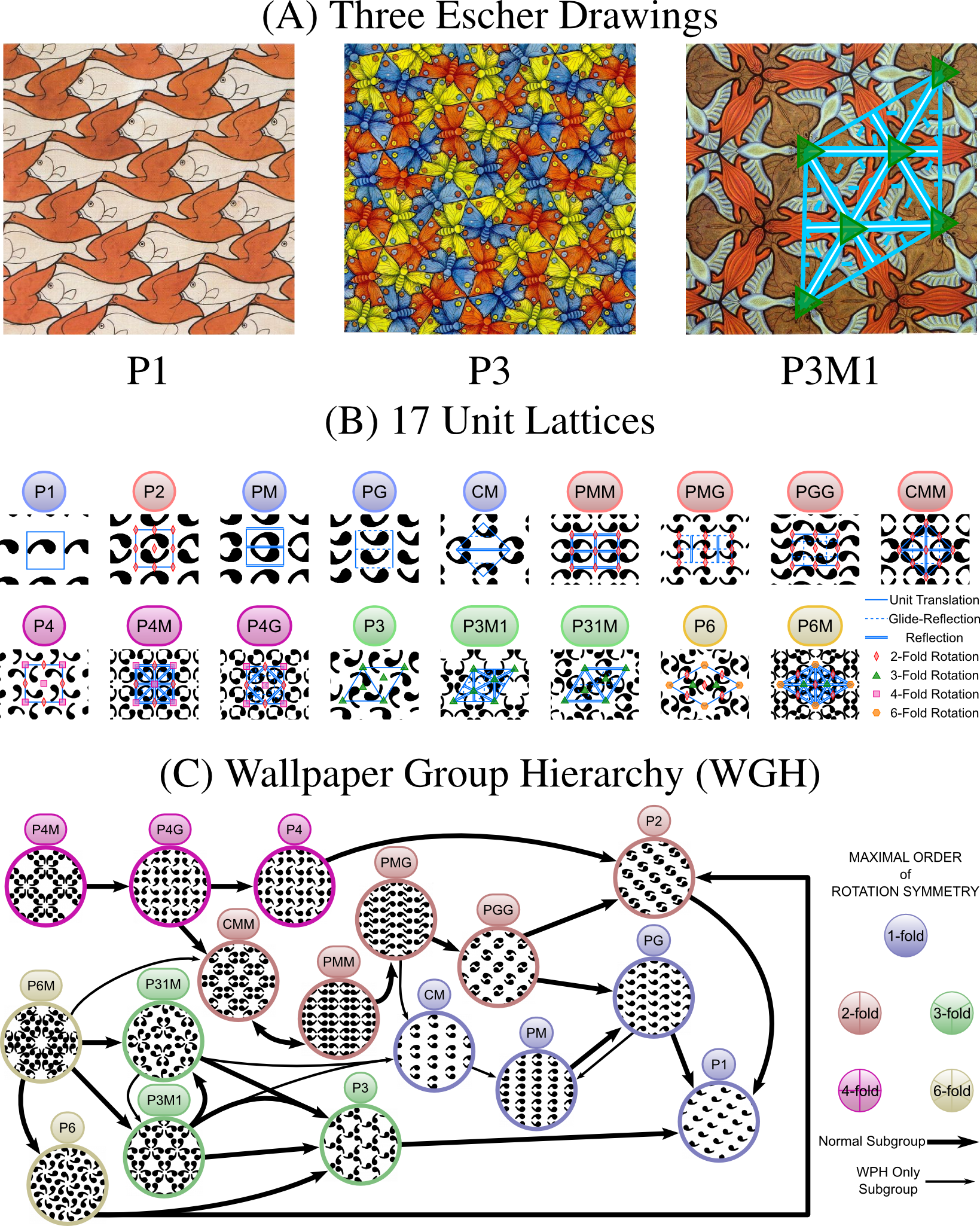}
	\vspace{-10pt}
	\caption{ 
		(A) Three drawings by M.C. Escher. 
		(B) {\bf Unit lattice}: The inner-structure for each wallpaper group, the unit lattice symmetries,
		originally described in \citet{schattschneider1978plane} using Wade's~\cite{wade1993crystal} comma visualizations. 
		(C) {\bf Wallpaper Group Hierarchy (WGH)}: 
		adapted from \citet{liu2001skewed,liu2010computational}, originally proven in \cite{coxeter1980generators}.
		Note: $A \rightarrow B$ means $B$ is a subgroup of $A$.
	}
	\label{fig:WP_Groups}\label{fig:WP_Hierarchy}
\end{figure}

Periodic patterns in $N$-dimensional Euclidean space are well characterized by group theory, namely, the {\em Crystallographic Groups} \cite{coxeter1980generators}. 
These groups form a 
{\bf complete} categorization for all possible periodic patterns in $R^N$;  
for a given $R^N$, there is a proven {\bf finite} set of $M_N$ distinct symmetry groups.
When $N=2$, $M_2=17$, the seventeen {\em wallpaper groups}.  
Figure~\ref{fig:WP_Groups} (A) 
demonstrates three M.~C.~Escher\footnote{17 June, 1898 to 27 March, 1972, a Dutch graphic artist who made mathematically inspired woodcuts, lithographs, and mezzotints.} drawings with corresponding wallpaper group indicated.
These 17 wallpaper groups are distinct (Figure~\ref{fig:WP_Groups} (B)) but not mutually-exclusive since they form a subgroup-hierarchy (Figure~\ref{fig:WP_Groups} (C)), called {\em Wallpaper Group Hierarchy} (WGH).
M.C. Escher is the namesake for this paper to honor his art, which contains many of the wallpaper patterns
(Figure \ref{fig:WP_Groups}(A)), and {\em EscherNet 101} indicates an initial effort on 
deep learning of wallpaper groups from synthesized wallpaper patterns.

Understanding periodic patterns has a direct relevance to texture perception  \cite{portilla_simoncelli_IJCV2000parametric,liu_etal2004SIGGRAPH,kohler2016representation}. 
Based on first principles, a computational framework for wallpaper group classification from a given image was proposed in early 2000 \cite{liu_etalPAMI2004}.
Little progress has been made on automatically learning transformation-invariant properties (symmetries) from periodic patterns directly. 
A fMRI/EEG study \cite{kohler2016representation} using wallpaper patterns as visual stimuli reveals that human neural responses are significantly affected by the highest order of rotation symmetry in wallpaper patterns. 
However, the visual stimuli in that study only included a subset, P1,P2,P3,P4,P6 (Figure \ref{fig:WP_Groups}(B,C)), from the 17 wallpaper groups. 

Using Convolutional Neural Network (CNN) as a computational model for machine perception and group theory as a principled taxonomy for class labels, we explore the feasibility of learning symmetry and symmetry group hierarchy (the WGH) directly from images of periodic patterns. 
Our contributions include: \\
(1) systematically training and testing a deep learning model for categorizing images of periodic patterns into their respective 17 wallpaper groups; \\
(2) providing unique observations on learning second-order invariants and class-hierarchies,  
through inspections of the learned filters at different levels of perception related to human vision. \\

\label{sec:images}
\begin{figure*}[!t]
	\centering
	\includegraphics[width=.98\linewidth]{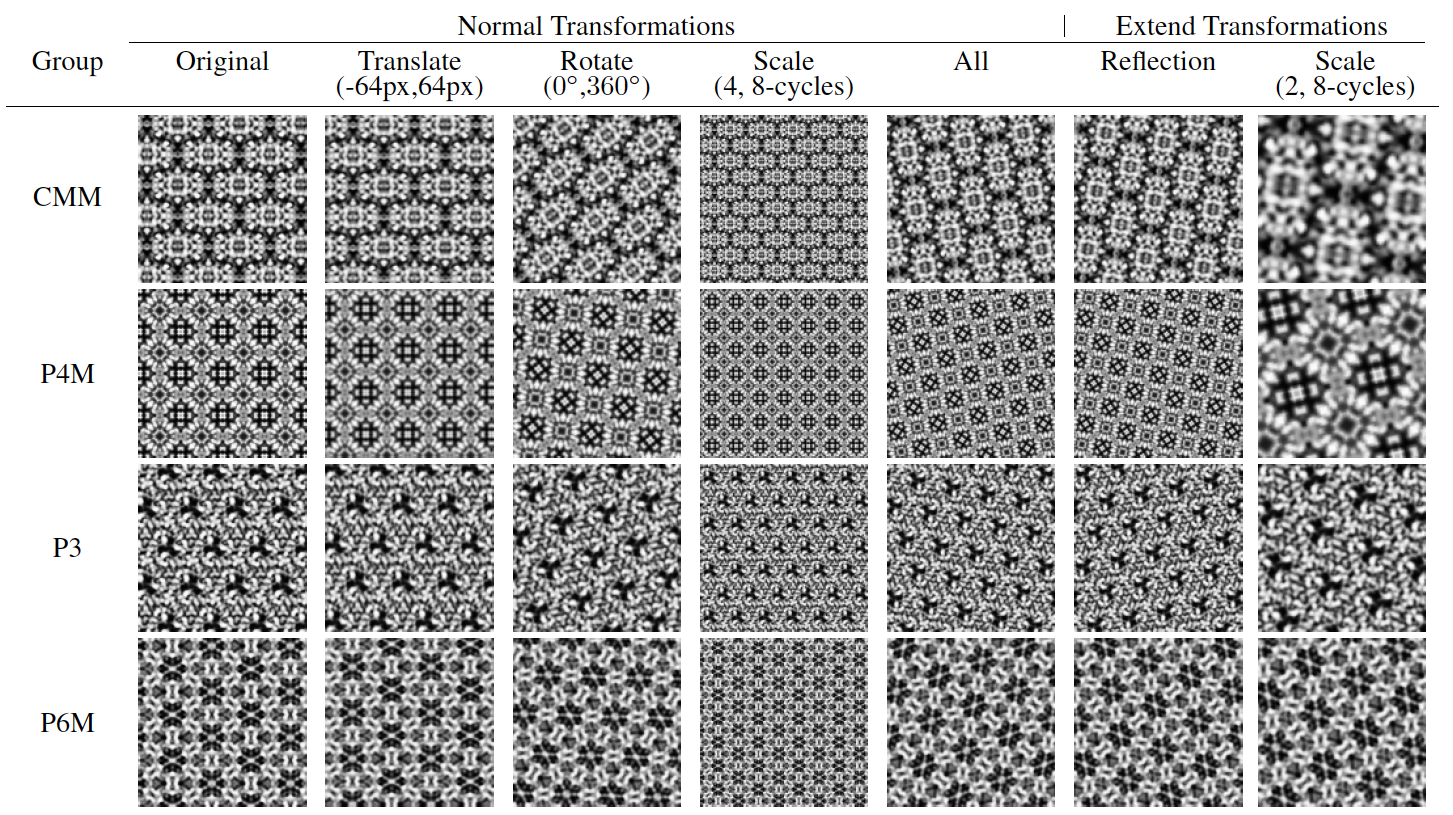}
	\setlength{\tabcolsep}{1pt}
	\renewcommand{\arraystretch}{.2}
	\resizebox{\linewidth}{!}{
	\begin{tabular}{*{8}{c}}
		 \midrule
	\# Train &  Original & Translate  & Rotate & Scale & All & Reflection & Scale  \\
	&& (-64px,64px) & ($0^\circ$,$360^\circ$) & (4, 8-cycles)  &  &  & (2, 8-cycles)\\
	\midrule
	 10
	 & $ 14.2\% \pm 6.1\% $
	 & $ 14.8\% \pm 5.4\% $
	 & $ 15.6\% \pm 6.3\% $
	 & $ 17.1\% \pm 9.5\% $
	 & $ 16.2\% \pm 8.4\% $
	 & $ 16.2\% \pm 8.3\% $
	 & $ 13.1\% \pm 7.4\% $
	 \\\midrule 
	 15
	 & $ 16.2\% \pm 10.8\% $
	 & $ 18.1\% \pm 7.5\% $
	 & $ 20.1\% \pm 7.6\% $
	 & $ 22.1\% \pm 10.9\% $
	 & $ 20.6\% \pm 9.6\% $
	 & $ 20.7\% \pm 9.7\% $
	 & $ 15.7\% \pm 9.2\% $
	 \\\midrule 
	 20
	 & $ 37.9\% \pm 16.6\% $
	 & $ 41.2\% \pm 12.6\% $
	 & $ 44.8\% \pm 10.7\% $
	 & $ 48.4\% \pm 12.4\% $
	 & $ 46.3\% \pm 11.7\% $
	 & $ 46.1\% \pm 11.5\% $
	 & $ 29.8\% \pm 10.5\% $
	 \\\midrule 
	 25
	 & $ 38.3\% \pm 19.3\% $
	 & $ 42.8\% \pm 14.3\% $
	 & $ 46.6\% \pm 12.6\% $
	 & $ 55.0\% \pm 11.5\% $
	 & $ 50.4\% \pm 12.0\% $
	 & $ 50.4\% \pm 12.0\% $
	 & $ 32.2\% \pm 11.4\% $
	 \\\midrule 
	 30
	 & $ 60.0\% \pm 21.0\% $
	 & $ 63.9\% \pm 17.5\% $
	 & $ 70.2\% \pm 15.1\% $
	 & $ 71.4\% \pm 12.5\% $
	 & $ 71.4\% \pm 14.2\% $
	 & $ 71.3\% \pm 14.3\% $
	 & $ 44.3\% \pm 11.5\% $
	 \\\midrule 
	 50
	 & $ 67.6\% \pm 22.6\% $
	 & $ 74.7\% \pm 14.4\% $
	 & $ 81.0\% \pm 11.2\% $
	 & $ 84.8\% \pm 10.4\% $
	 & $ 84.0\% \pm 10.2\% $
	 & $ 84.0\% \pm 10.3\% $
	 & $ 52.0\% \pm 11.4\% $
	 \\\midrule 
	 75
	 & $ 76.5\% \pm 20.8\% $
	 & $ 84.7\% \pm 12.5\% $
	 & $ 89.6\% \pm 9.1\% $
	 & $ 91.3\% \pm 8.0\% $
	 & $ 91.5\% \pm 7.7\% $
	 & $ 91.5\% \pm 7.7\% $
	 & $ 57.1\% \pm 11.8\% $
	 \\\midrule 
	 100
	 & $ 71.6\% \pm 26.3\% $
	 & $ 83.6\% \pm 9.8\% $
	 & $ 91.7\% \pm 5.7\% $
	 & $ 92.9\% \pm 4.7\% $
	 & $ 93.6\% \pm 5.1\% $
	 & $ 93.5\% \pm 5.1\% $
	 & $ 59.1\% \pm 10.6\% $ 
	 \\\midrule 
	 600
	 & $ 74.5\% \pm 23.5\% $
	 & $ 87.9\% \pm 8.7\% $
	 & $ 96.5\% \pm 3.1\% $
	 & $ 95.2\% \pm 3.6\% $
	 & $ 97.5\% \pm 2.2\% $
	 & $ 97.4\% \pm 2.4\% $
	 & $ 60.4\% \pm 9.5\% $
	 \\\midrule 
	 6k
	 & $ 83.4\% \pm 17.4\% $
	 & $ 93.3\% \pm 5.9\% $
	 & $ 98.3\% \pm 2.1\% $
	 & $ 97.8\% \pm 2.1\% $
	 & $ 98.7\% \pm 1.7\% $
	 & $ 98.7\% \pm 1.8\% $
	 & $ 62.3\% \pm 10.1\% $
	 \\\midrule 
	 48k
	 & $ 79.9\% \pm 19.3\% $
	 & $ 92.7\% \pm 5.7\% $
	 & $ 98.3\% \pm 2.1\% $
	 & $ 97.9\% \pm 2.0\% $
	 & $ 98.9\% \pm 1.6\% $
	 & $ 98.8\% \pm 1.7\% $
	 & $ 62.6\% \pm 10.9\% $
	 \\\midrule 
	 U-Method
	 &
	 &
	 &
	 & 
	 & $ 96.4\% \pm 8.3\%$
	 &
	 &
	 \\\bottomrule 
	\end{tabular}
	}
	\caption{\textbf{Top Images:} Sample wallpaper pattern image-input to EscherNet: 
		EscherNet trained on 48k WPIs per group, each training/testing image is randomly translated, rotated, and scaled.  Scale is shown at the extreme of  (8-cycles).  We also tested some of the classifications on augmentations not trained on such as an expanded scale which scales from 2 to 8 cycles (having the area of 2 cycles of unit lattices in the image) and reflection.  
		\textbf{Bottom Table:} The accuracy and standard deviation between the group means for the networks trained with different amounts of training data.  These networks are tested on 12,000 WPIs per group with original, translation, rotation, and scaling augmentations added.  The EscherNet is not affected reflection but the expanded scale makes the performance dip.  These addition transformations were tested while allowing for a random scale and translation.  The U-method results are also shown here.  \label{tab:aug classification matrix}  \label{tab:expanded transformations}  \label{tab:class accuracies} \label{fig:data}}
        \vspace{-15pt}
\end{figure*}

\section{Related Work}
\label{Related Work}

\noindent{\bf Symmetry Group Detection from Images}  

Deep learning is a state-of-the-art machine learning mechanism that has been used successfully in many 
computer vision tasks, particularly for object classification~\cite{krizhevsky2012imagenet,zeilerECCV2014visualizing,simonyan2013deep,yosinski2015understanding,szegedy2013intriguing,jaderbergECCV2014reading,agrawal14analyzing}.  However, how deep learning methods learn second order knowledge such as the invariant-under-automorphism (symmetry) properties within the original data is less explored.
Related to local-reflection symmetries, some learning-based medial-axes or skeleton symmetry detection works have been presented in recent years \cite{tsogkas2012learning,shen2016object,teo2015detection}.  \citet{tsogkas2012learning} use multiple instance learning and \citet{teo2015detection} use a structured random forest to create a symmetry probability map over the image.   \citet{shen2016object} create a deep network to learn scale dependent labels which are fused together to create the probability map.  
These approaches learn the medial axis of an object thus are object-dependent (but not object class dependent).   
Recently, \citet{funk_liu_2017_beyond} show the results of learning (CNN) from human labeled reflection and rotation symmetries on natural images to perceive beyond-planar symmetries that are object and object-class independent.  

Classic symmetry detection algorithms from images are mostly based on the mathematical definition of symmetry, followed by some type of clustering method \cite{liu2010computational,liu_etalPAMI2004, loy2006detecting, lee2012curved, lee2008rotation, lee2009curved,lee2009slewed,hays2006discovering,park2010translation,park2008deformed}.
Symmetry group detection algorithms on real images are rare; some examples include wallpaper group classification under Euclidean \cite{liu_etalPAMI2004} and affine \cite{liu2001skewed} transformations; and
discrete and continuous rotation symmetry groups under affine deformation \cite{lee2009slewed}.
See \cite{liu2010computational} for a survey.     

\noindent{\bf Transformation Groups applied to CNNs}

\citet{gens2014deep} define a network invariant to high-dimensional symmetries by pooling many rotations, translations, and scales at each layer.  \citet{cohen2016group,cohen_welling_ICLR2017_steerable} shift the filters during the convolution by the transformations of a particular group. 
These exciting recent works \cite{cohen_welling_ICLR2017_steerable, cohen2016group,gens2014deep} use transformation groups to generate transformed versions of the input data and activations within a CNN network (regardless whether the input data itself is invariant under any automorphism), thus enhancing CNN performance. In contrast, our work focuses on recognizing the inherent Euclidean symmetries in a periodic pattern image and on characterizing each input pattern by its corresponding wallpaper group. 

\noindent{\bf Visualization}
	
For the most part, the mechanism of deep learning remains a black box in many respects, such as an understanding of transformation invariance within the deep networks.  Recent work in understanding networks have focused on either individual image network response or overall dataset dependent analysis with visualizations of specific aspects of the network~\cite{zeilerECCV2014visualizing,simonyan2013deep,yosinski2015understanding,szegedy2013intriguing,agrawal14analyzing}. 
We train a deep neural network on a well-defined problem of classifying wallpaper patterns to their proven correct wallpaper groups, in order to gain an understanding of where and what the network is learning.  
	
\citet{agrawal14analyzing,girshick2014rich} use a quantitative approach to understanding what is learned at each CNN layer based on the entropy of the filters.  The lower the sum of the normalized entropy of label for increasing amounts of filters, the more discriminative the filter.  The entropy is normalized by the number of filters for each layer but the ranges are not normalized.
There have been a number of other attempts at visualizing what network's are learning.  Many approaches use the network's built-in back propagation to visualize the network.  \citet{zeilerECCV2014visualizing} use a deconvolution-based method of back propagation where a single high activation is back-propagated to see what is activating neuron in the image.  \citet{simonyan2013deep} use a similar method but from the final layer (before the softmax normalization). 
Another common visualization approach is to find or synthesize an image though gradient descent~\cite{simonyan2013deep,yosinski2015understanding,nguyen2015deep,mahendran_vedaldi_IJCV2016_visualizing}.  They try to determine what is activating the each neuron  synthesizing an image via increasing the neuron's activations or matching another image's activation under different regularizations.    
	Class Activation Mapping (CAM) by \citet{zhou_etal_CVPR2016learning} have tried to combine information on the activations at different layers for a particular class.  Zhou \textit{et al.} weights each filter's activation by the activation's impact on the final classification score and pools them together to visualize each pixel's importance in the original image.  Grad-CAM by \citet{selvaraju2016grad} extends this by being able to do it in one backwards propagation step.  %\citet{zhang2016top} use excitation back propagation which only back propagates the positive gradients as well as negating the gradients relevant to the other classes.  
	T-SNE reductions \cite{van2008visualizing,van2014accelerating} have been shown to help visualize the output of the last fully connected layer~\cite{joulin2016learning}.  This is a useful tool for visualizing the approximate distance between classes; however, it has the limitation of not being able to truly show the distances between classes because of the reduction of dimension~\cite{joulin2016learning}.

Finally, \citet{jaderbergNIPS2014sythetic} train a network using synthetic data to extend their training set to enhance natural scene text recognition.  This shows the value of using synthetic data in training networks. 
There have been a number of other attempts at visualizing what the networks are learning.  \citet{zeilerECCV2014visualizing} use a deconvolutional approach which mimics backward propagating a high activation back through the network.  
\citet{simonyan2013deep} use a similar method to \citet{zeilerECCV2014visualizing} but from the final layer (before the softmax normalization).  The only difference between the two methods is within the relu implementation.  In \citet{zeilerECCV2014visualizing} the relu is done after the deconvolution (which is switching the order of the original network) rather than on the data from the previous layer.  \citet{simonyan2013deep} only visualized the last fully connected layer.  
Another common visualization approach is to find or synthesize an image though gradient descent.\cite{simonyan2013deep,yosinski2015understanding,nguyen2015deep}  In \citet{simonyan2013deep}, a target image is chosen and the euclidean loss between the target image's activation and the randomly generated starting image is optimized to try and reproduce the target's activation.  This method can identify some invariance of the neuron by looking at the differences between the optimized image and the target image.  In order to try and make the images more recognizable, \citet{simonyan2013deep} used $ \ell_2 $ regularization on the activations.  These images seem to only incorporate high frequency information within them.  Others have expanded on their work and use more complex regularizations such as \cite{mahendran2015understanding,yosinski2015understanding}.  Others have extended this idea by propagating the final layer activation for another class \cite{szegedy2013intriguing} and using a genetic algorithm to produce almost unrecognizable pictures which are classified to have extremely high confidence \cite{nguyen2015deep}.  These approaches only show information about a single image prior and a layer or neuron and not the general layer invariance between the images.

\section{Our Approach}
\label{sec:Methods}
We aim to find out (1) whether and (2) how an artificial neural network may learn {\em symmetry groups} and their relations from input data labeled by symmetry groups.
Since symmetry is a non-local, abstract concept, the images $S_i$ belonging to the same symmetry group class $G$ may not share any first-order visual cues such as edges, shape contours or intensities.  Instead, the only thing they share is a set of closed transformations $G$ such that $g(S_i) = S_i$ for all $g \in G, i = 1 ... N$ (Figure~\ref{fig:data}).
Utilizing an effective wallpaper-pattern synthesis algorithm~\cite{kohler2016representation} that starts from a noise patch and follows a specific set of group generators \cite{coxeter1980generators}, we 
train a convolutional neural network for wallpaper group classification under different training/testing conditions.
On average, a nearly perfect classification rate on the 17 wallpaper groups is achieved (Table \ref{tab:aug classification matrix}). 
Even more challenging perhaps is to understand {\em where} in the layers of a CNN the individual symmetries and their particular combinations within each symmetry group are being learned. 
To this end, we 
apply different visualization tools to understand the filters and their relations.

Wallpaper pattern images (WPIs) in this study are synthesized images, generated based on a variant of a wallpaper pattern synthesis algorithm 
\cite{kohler2016representation}.
The generator uses histogram equalization to avoid any intensity differences among the images.  
The statistics of image intensity of each WPI are verified on all training images with an intensity mean of $0.5 \pm 0.00037$ for all 17 wallpaper groups respectively.
The WPIs are generated from a random patch of grayscale noise.  
Each WPI has 256x256 pixel, 1-channel, with 8x8 unit-lattices.  
For training,  60,000 WPIs are created per wallpaper group, and randomly augmented with rotation, scale and translation (Figure \ref{fig:data}).
The images are translated by a max range of [-64,64] pixels, rotated between [$ 0^\circ $, $ 360^\circ $], 
scaled between 100\% and 200\%  
from 256x256 pixels,\footnote{
	The original image is not scale lower than 100\% so the entire input into the network contains the wallpaper pattern.}, 
and cropped to 128x128 pixels (the input size of the network).
\footnote{ The original image is cropped to 181x181 when comparing against other augmentations since 181 pixels is the minimum allowable size after rotation: $ 256*\cos(45^\circ) \approx 181 $.}

\noindent{\bf EscherNet 101 Training and Testing Design}
EscherNet 101 is a CNN consisting of 4-layers.  The first 2 layers are two repetitions of Conv-ReLU-Conv-ReLU-Pool with 32 7x7 filters for the first two Conv layers (conv1 and conv2) and with 64 3x3 filters for the others (conv4 and conv5).  The max pool layers are 2x2 with a stride of 2.  These are followed by two fully connected layers with 512 and 17 neurons, respectively, with a ReLU between them, and a final softmax layer.    We choose a small network with few parameters because it is easy to analyze while still being able to achieve a high classification rate. 
We train the EscherNet using a training set of 
48,000/6,000/600 WPIs per group respectively.   
The test set is the same and consists of 12,000 WPIs per group (204,000 total WPIs).  
Randomized augmentations are applied to the testing set.

EscherNet is trained from scratch using Keras~\cite{chollet2015keras} with a backend of Theano~\cite{bergstraSciPy2010} optimizing by Stochastic Gradient Descent and minimizing the cross-entropy loss. The learning rate is 0.01, 0.005, and 0.001 (reduced when converging at a loss rate), and a batch size of 250 is used during training.    
Loading and processing the images with the augmentation takes 3ms per image. 

\begin{figure*}[!ht]
	\centering
	\includegraphics[width=.8\linewidth]{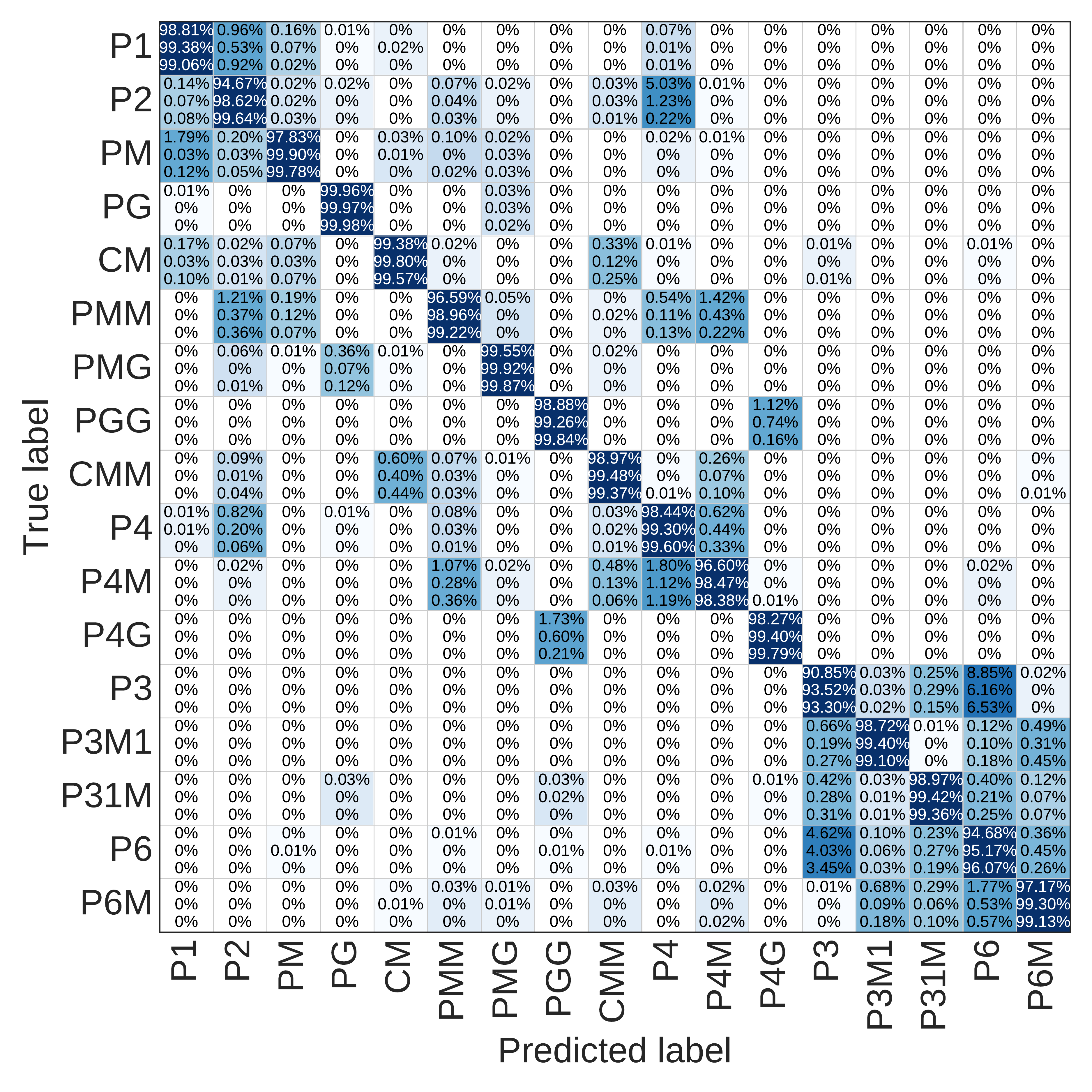}
	\caption{The confusion matrix for EscherNet 101 tested on all data transformations on 12,000 WPIs per group.  
		Within each cell, three rates are shown for the networks trained on 600 (top), 6,000 and 48,000 WPIs per group respectively.   
		The diagonal cells contain the classification rates of the wallpaper group (p3 and p6 are the lowest two) and the off-diagonal cells $i,j$ show the error rate of classifying wallpaper group $i$ to group $j$ (p3 classified to p6 is the highest one) .
	}
	\label{fig:confusion matrix}
\end{figure*}

\noindent{\bf Experimental Results}
\label{Results} 
Table \ref{tab:aug classification matrix} and Figure~\ref{fig:confusion matrix} demonstrate the resulting classification rates and the corresponding confusion matrix.  
The similar classification rates suggest that over 600 WPIs/group, the true positive rates seem to be insensitive to the training set size.

\section{Is the Network Learning Symmetry}
One issue that needs to be addressed is if the network is learning symmetry or is it learning something else.  We tackle this issue in two ways: (1) We show that when we lower the amount of training data, the network gets confused between symmetry groups which contain common symmetries and (2) we compare against classifying the WPIs using bag of words~\cite{csurka2004visual} with SURF descriptors~\cite{bay2008speeded} as features to see if the WPIs can be learned through only pixel-based descriptor features and, separately, on the Fourier Coefficients to see if they can be classified based on the periodicity of the patterns alone.

\subsection{Limited the Training Dataset}
How little data is needed for the EscherNet 101 to classify the wallpaper patterns can shed light into what the network is learning.  The less amount of data needed to accurately classify the patterns, the more likely EscherNet is classifying symmetry rather than just memorizing the look of certain patterns since there is much less variation in the look of the patterns as the amount of training data is reduced.  Also, focusing on where EscherNet becomes confused provides more information on what symmetries are being learned.  

\begin{figure}
	\centering
	\resizebox{\linewidth}{!}{
	\begin{tabular}{cc}
		10 WPIs per group & 15 WPIs per group \\
		\includegraphics[width=0.45\linewidth]{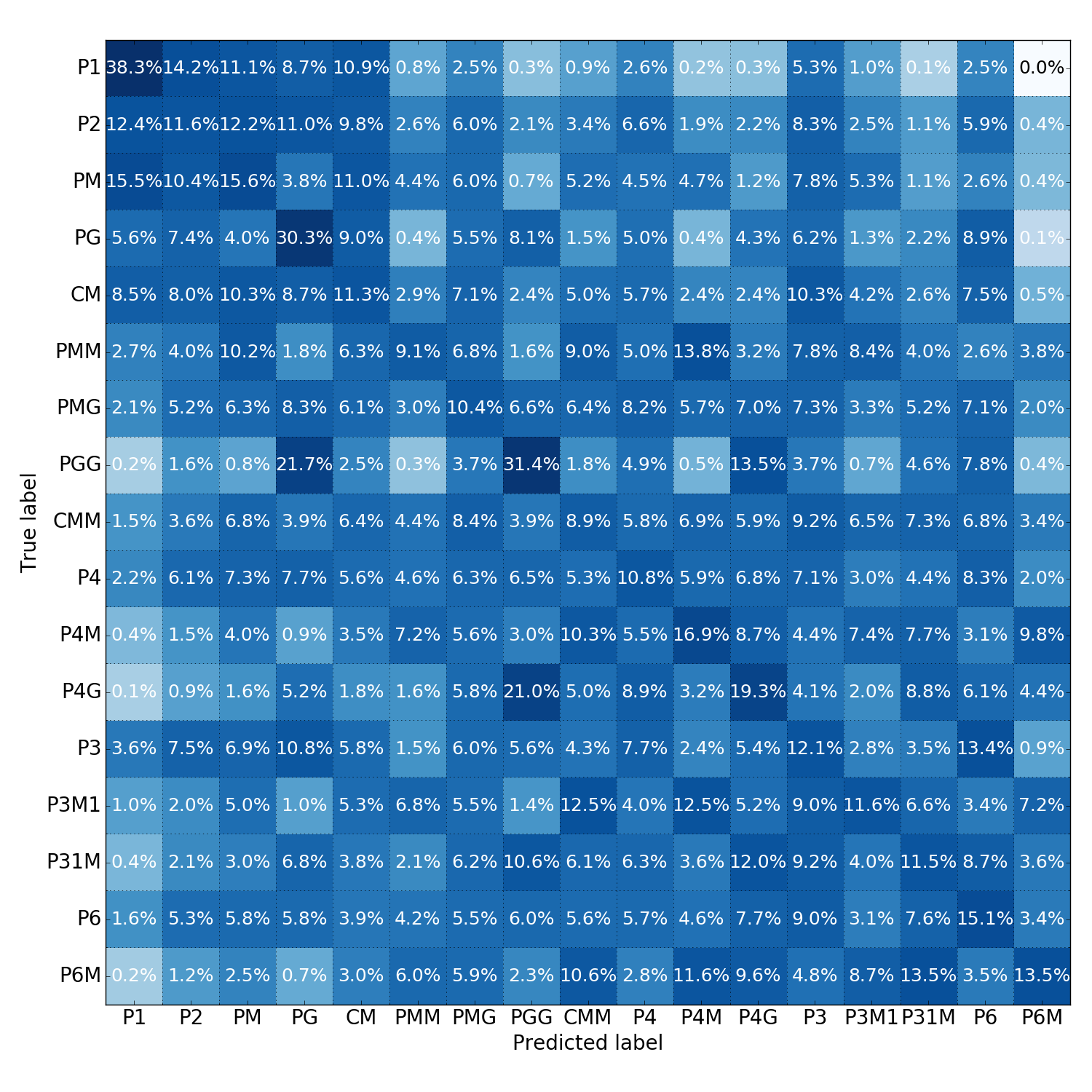} &
		\includegraphics[width=0.45\linewidth]{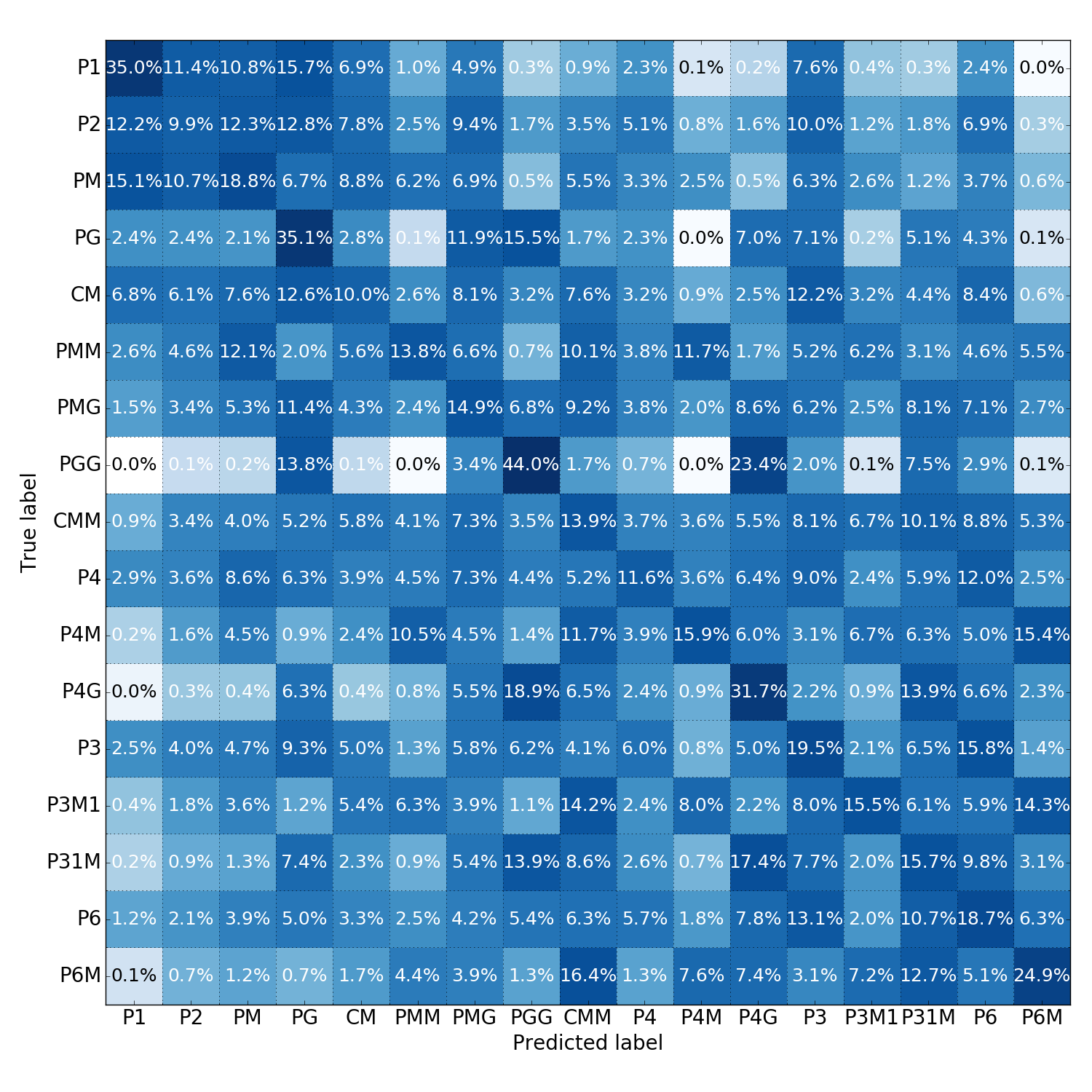}\\
		20 WPIs per group & 25 WPIs per group \\
		\includegraphics[width=0.45\linewidth]{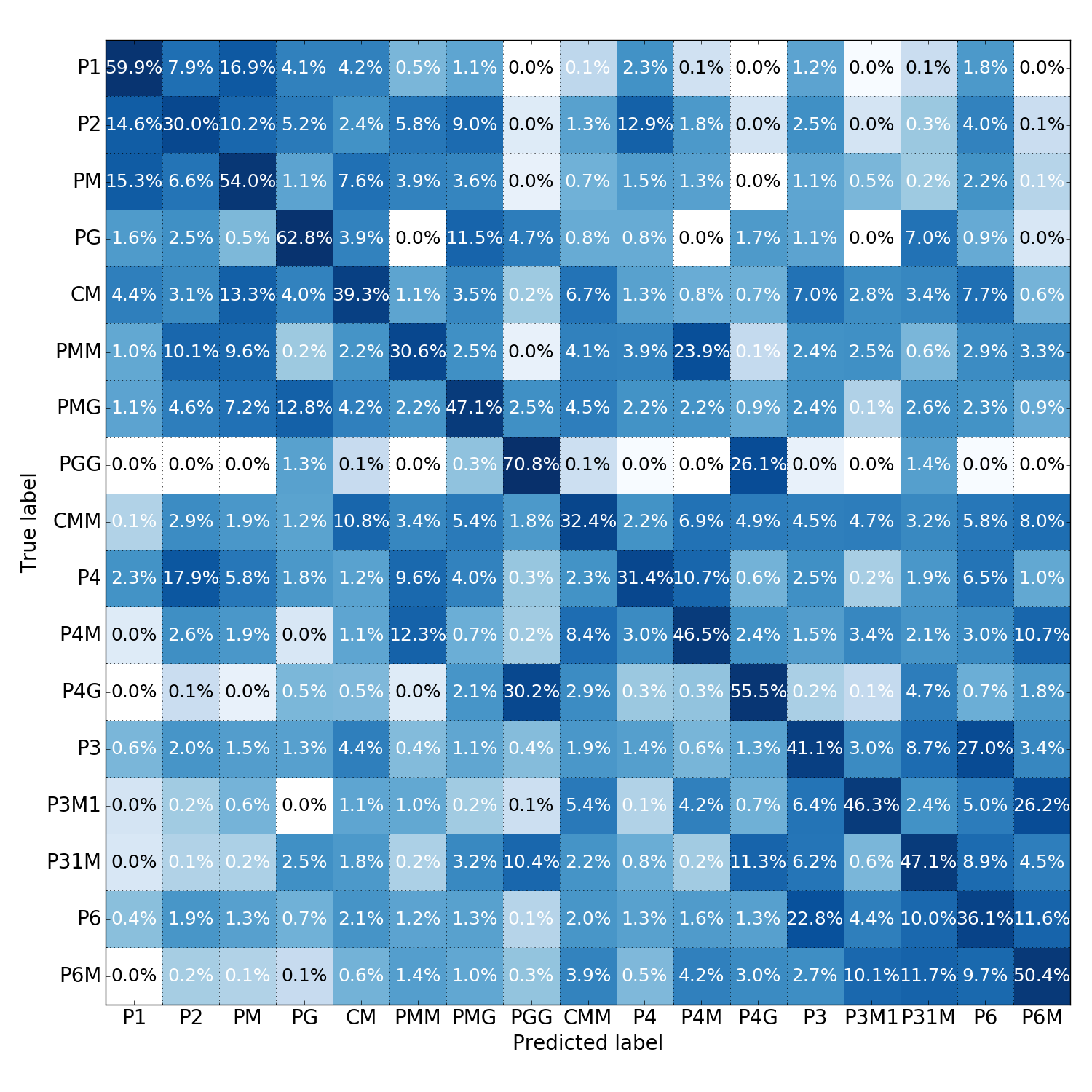} &
		\includegraphics[width=0.45\linewidth]{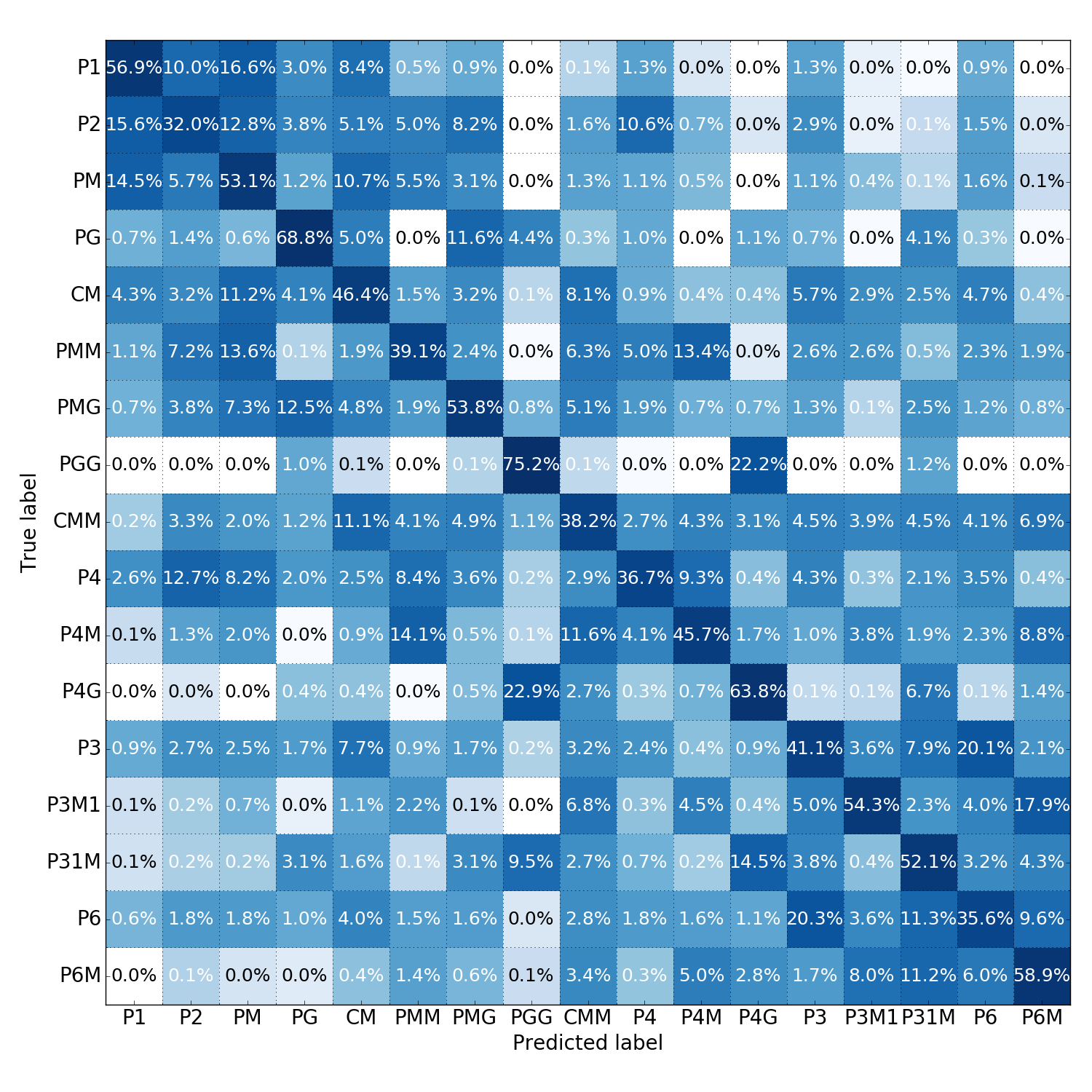}\\
		30 WPIs per group & 50 WPIs per group \\
		\includegraphics[width=0.45\linewidth]{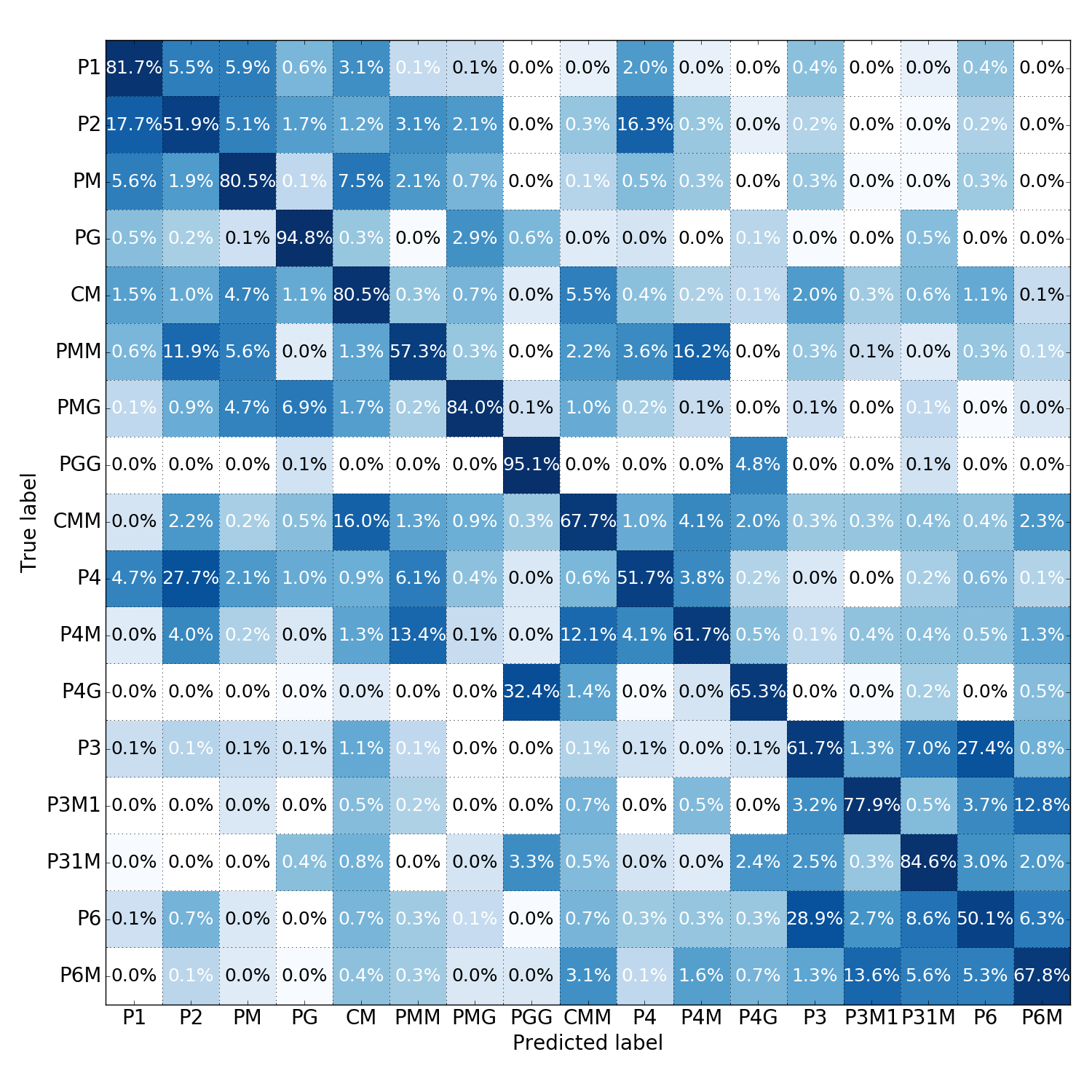} &
		\includegraphics[width=0.45\linewidth]{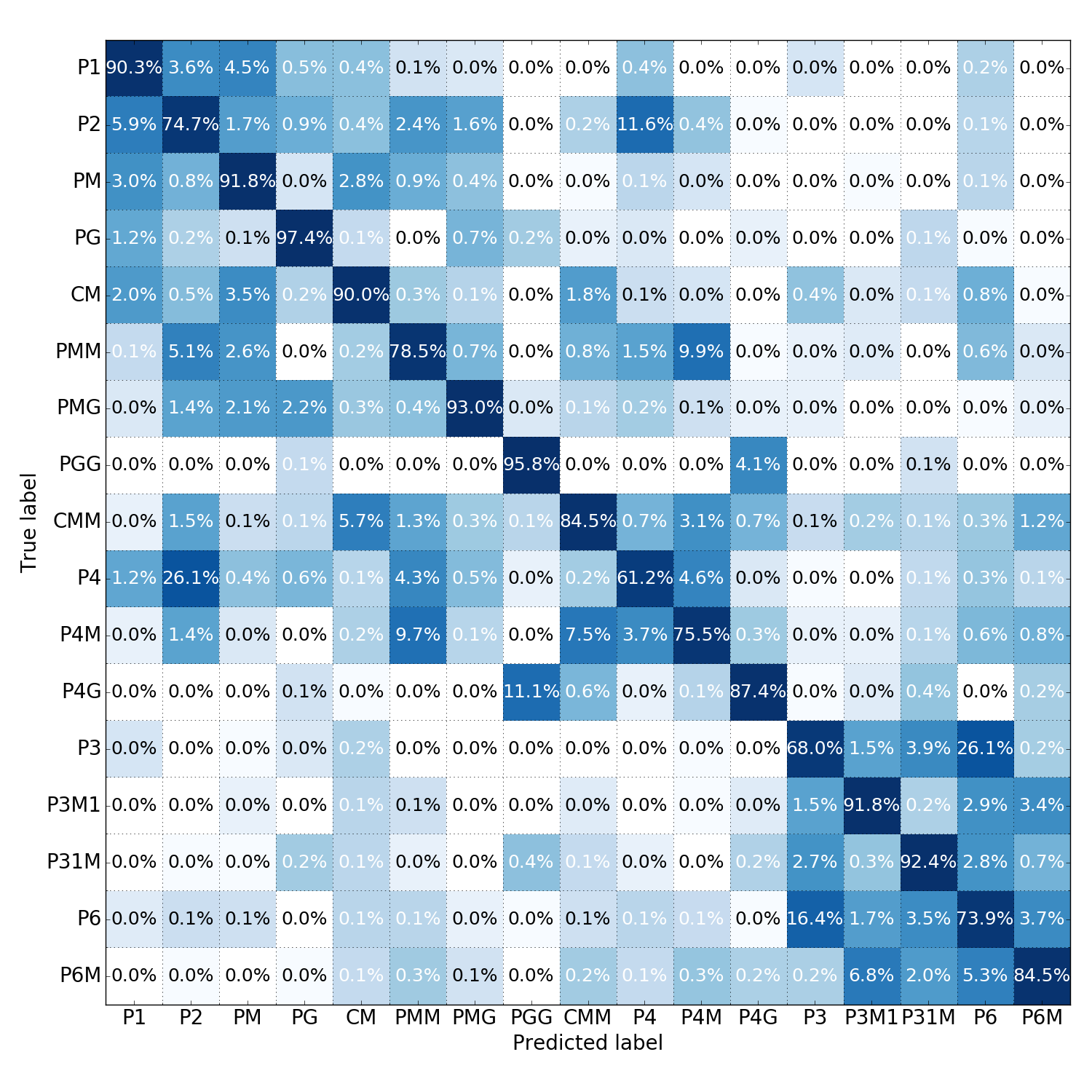}\\
		75 WPIs per group & 100 WPIs per group \\
		\includegraphics[width=0.45\linewidth]{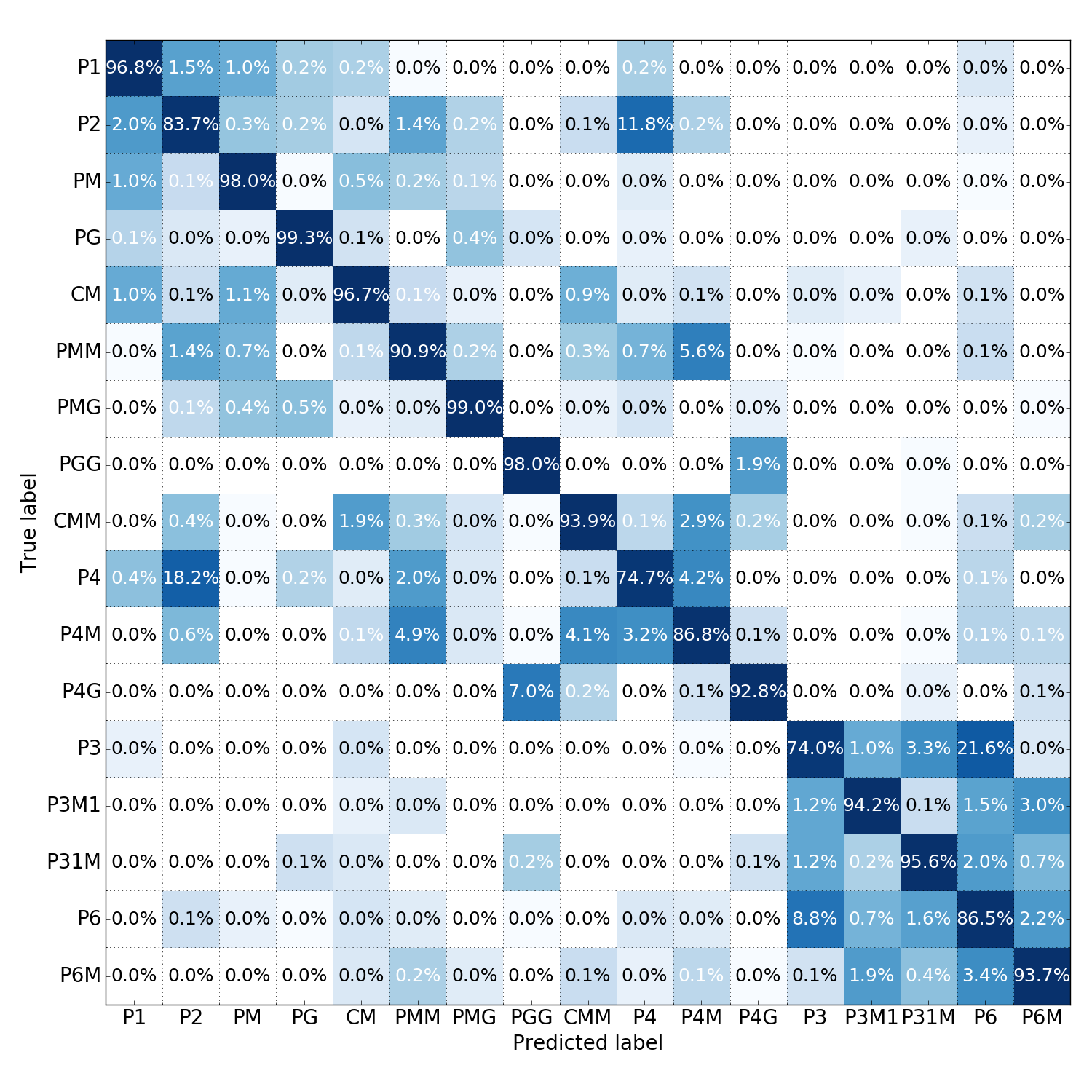} &
		\includegraphics[width=0.45\linewidth]{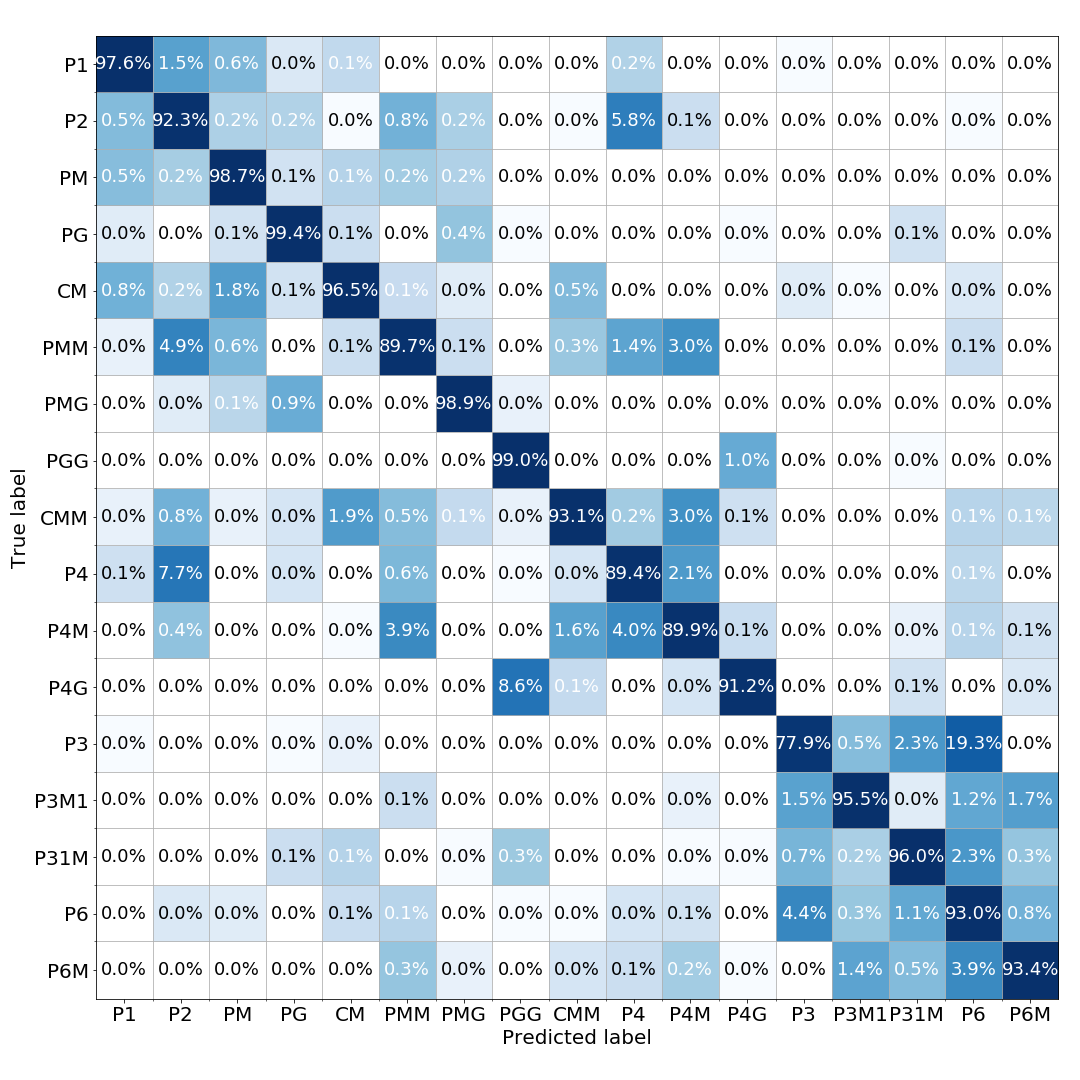}
	\end{tabular}
	}
	\caption{Confusion Matrices for EscherNet trained on between 10 and 100 WPIs per group.  Classification accuracies are shown in Table \ref{tab:class accuracies}.  The groups which are most confused share many symmetries (such as PG$\leftrightarrow$PGG for 10 WPIs per group and P3$\leftrightarrow$P6 for 100 WPIs per group) and tend to mostly differ in rotational symmetry.}
	\label{fig:cm small}
\end{figure}

We limit the training set down to a very small number of WPIs per group by starting from 10 patterns per group and progressively increasing to 100 WPIs per group while training from scratch for each amount of training data (Table~\ref{tab:class accuracies}).  The accuracy versus the amount of training data seems to have a logarithmic relationship similar to other deep learning problems~\cite{sun2017revisiting}.  EscherNet only takes 75 images per group (1,275 total images) to obtain a classification rate $ >90\% $ from a random initialization. 
The groups which are confused between the wallpaper groups share similar symmetries (confusion matrices: Figure~\ref{fig:cm small}) such as the maximum confusions between PG$\leftrightarrow$PGG for 10 WPIs per group and P3$\leftrightarrow$P6 for 100 WPIs per group, mainly differing in their rotational fold.  

\subsection{Classifying with Image Features} \label{sec:Image Features}
To show that the symmetries are being learned instead of just ``remembering" typical image features, we classify the data with an off-the-self image category classifier using bag-of-words~\cite{csurka2004visual} and Surf Descriptors~\cite{bay2008speeded}.  While there is some visual similarity in the patterns for the same group, if this is all EscherNet needs to classify the patterns then this approach should get a comparable accuracy.  

\begin{figure*}[ht]
	\centering	
	\resizebox{\linewidth}{!}{
	\begin{tabular}{cc}
		BoW SURF Image Features & Fourier Coefficient Features \\
		\includegraphics[width=0.49\linewidth]{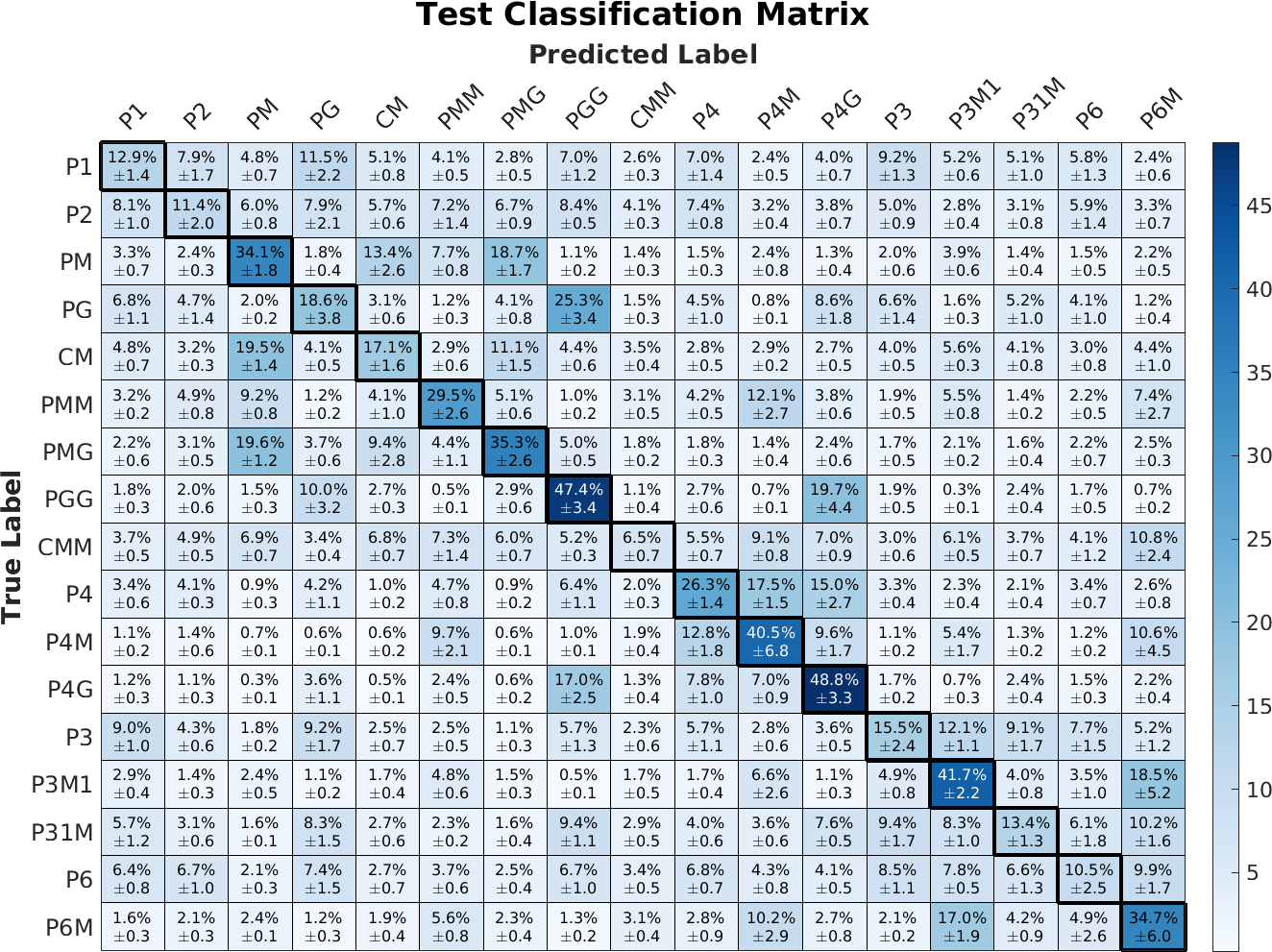} 		&
		\includegraphics[width=0.49\linewidth]{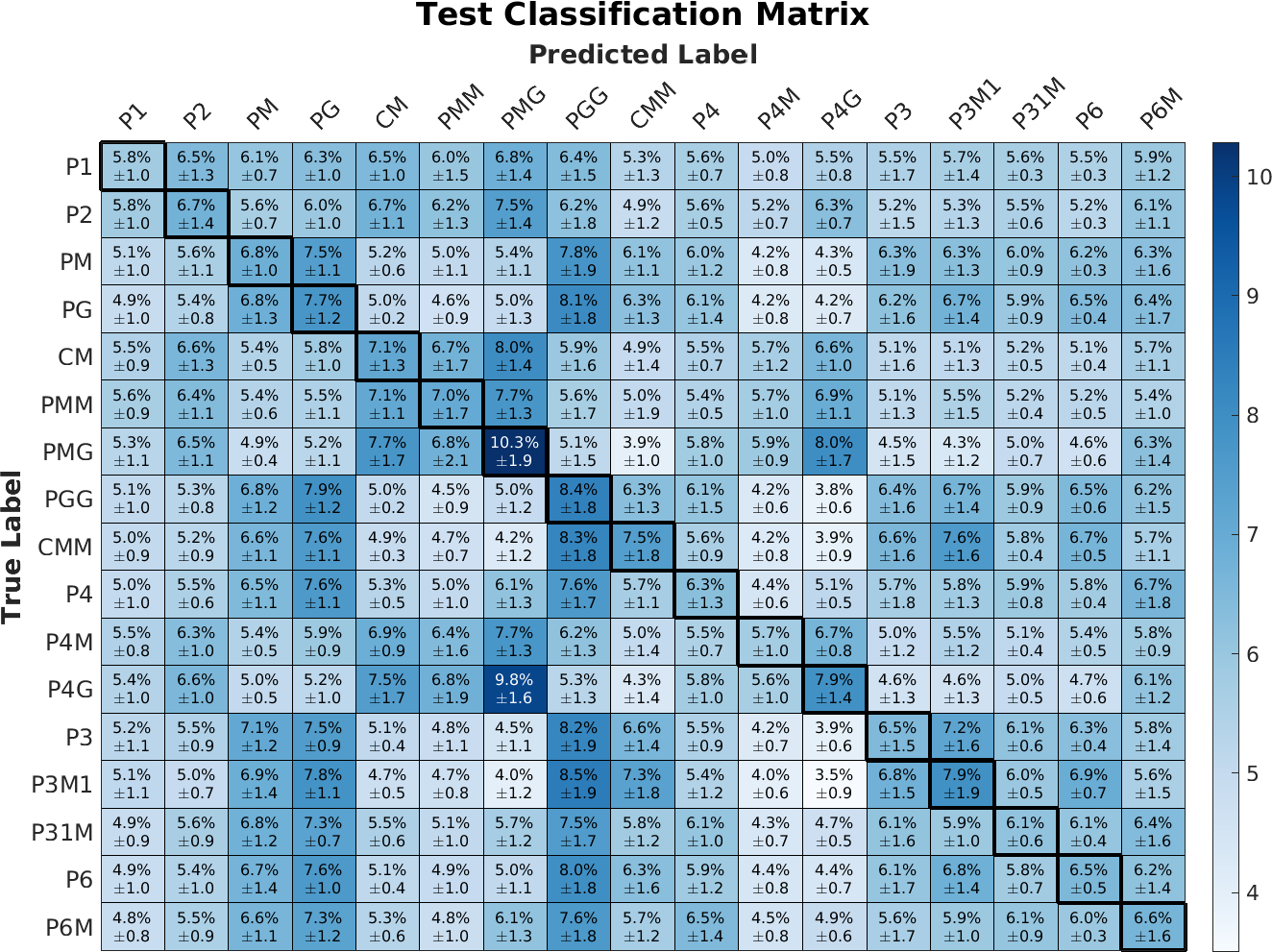} \\
        EscherNet-SVM & U-Method\\
		\includegraphics[width=0.49\linewidth]{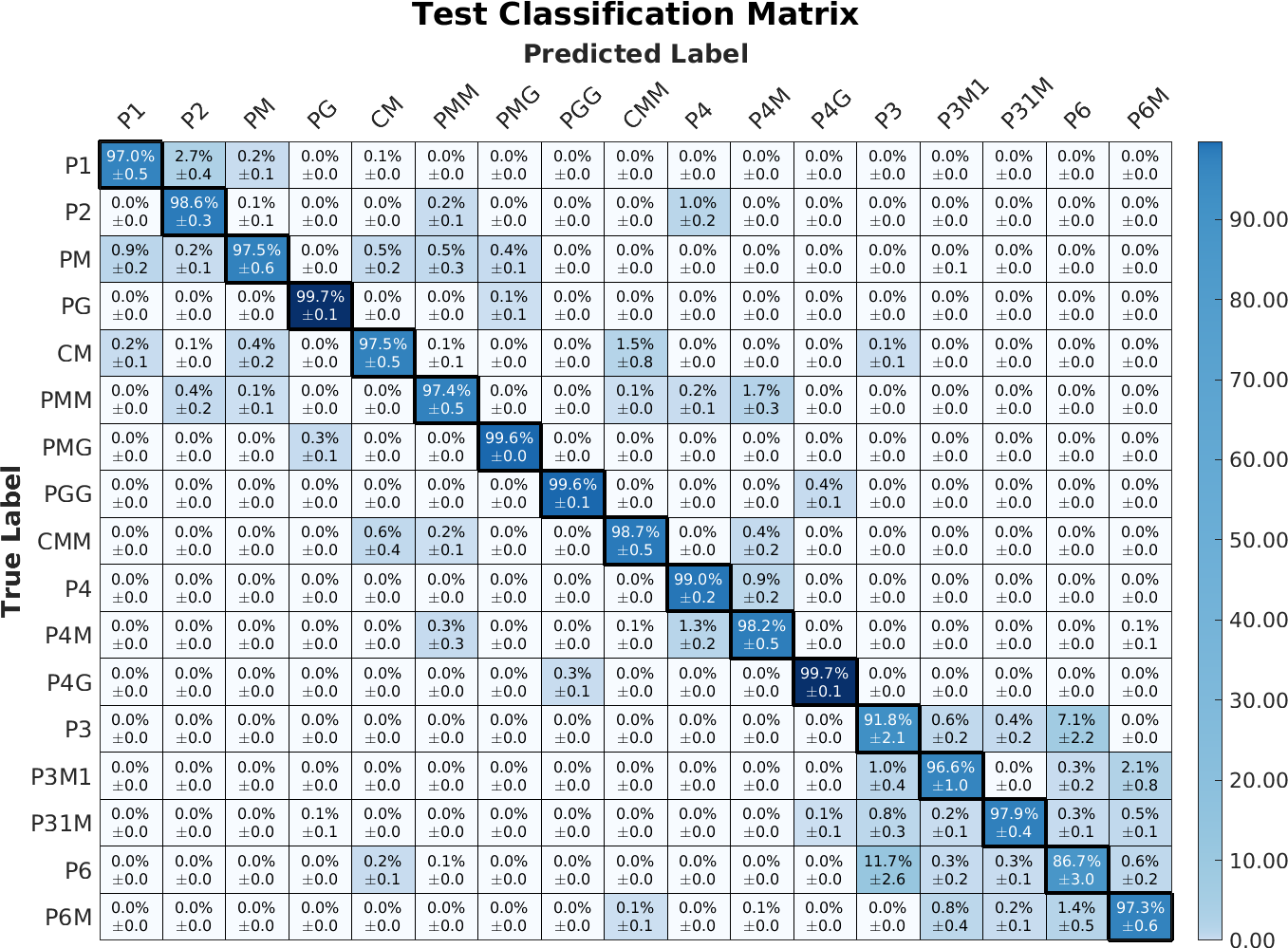}  & 
        	\includegraphics[width=.49\linewidth]{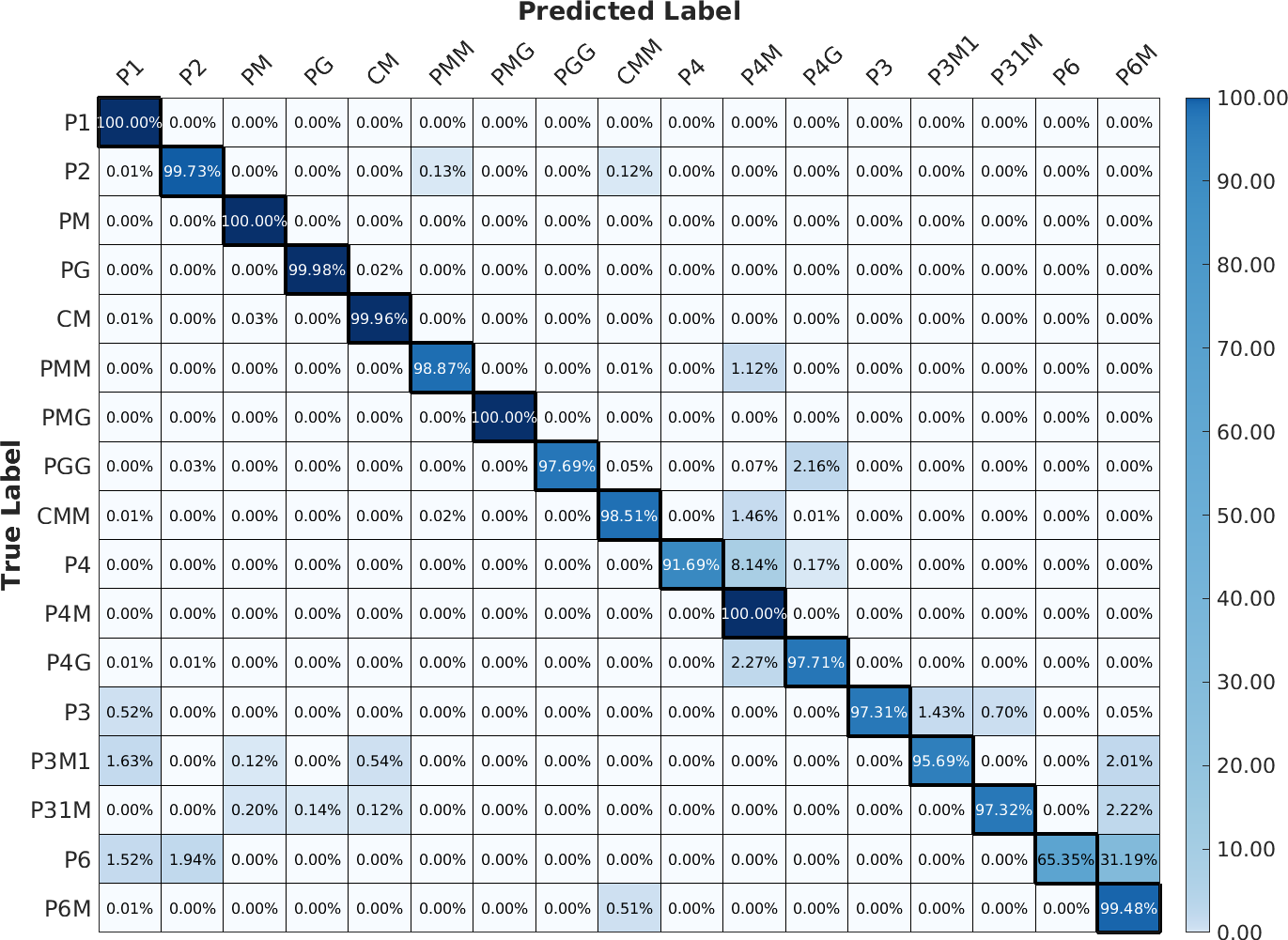} \\      	
        \end{tabular}
	}
	\caption{Confusion Matrices for experiments of classifying the the WPIs with 100 WPI per group of training images using bag-of-words SURF features, using Fourier Coefficients, and the last fully connected layer of the EscherNet trained on 48k as well as the U-method (unsupervised classifier) from \cite{liu_etalPAMI2004}.  These heatmaps are the mean and std between 5 classifications using 100 randomly chosen WPIs per group for training.  The main diagonal is highlighted to show the accuracy rates for each wallpaper group.  Classifying with BoW SURF features obtains a classification rate accuracy of $26.1\%\pm13.9$, Fourier Coefficient features obtained $ 7.1\% \pm 1.1\% $, the last layer of EscherNet obtained $97.22\% \pm 3.29\%$, and the U-Method obtrained $ 96.4\% \pm 8.3\%$. }
	\label{fig:avr results 12000} \label{fig:deterministic CM}
\end{figure*}

We use a training set of 100 WPIs per group (1,700 total) which are preprocessed the exact same way as EscherNet (with all augmentations).  Features from the WPIs are extracted using SURF and are clustered using a bag-of-words (BoW) with a vocabulary of 500.  These features are classified using a multiclass error correcting output codes (ECOC) classifiers~\cite{allwein2000reducing,escalera2009separability,escalera2010decoding} with Gaussian SVM learners~\cite{cristianini2000introduction}.  The multiclass classifier is then tested against the same test set of WPIs as EscherNet (12,000 WPIs per group).  We train the classifier 5 times with 100 randomly chosen WPIs per group from the training set and use the average between these classifications (Figure~\ref{fig:avr results 12000}).  

The classifier obtains an accuracy of $26.1\%\pm13.9$ which, while better than random chance, is nowhere near the $ 93.6\% \pm 5.2\% $ EscherNet obtained with the same amount of training WPIs.  This, however, does not eliminate the possibility that EscherNet might be using some image features for classification.  This only shows that using image features alone are most likely insignificant for learning to classify the wallpaper groups. 

\COMMENT{
\begin{figure}[ht]
	\centering	
	\includegraphics[width=.9\linewidth]{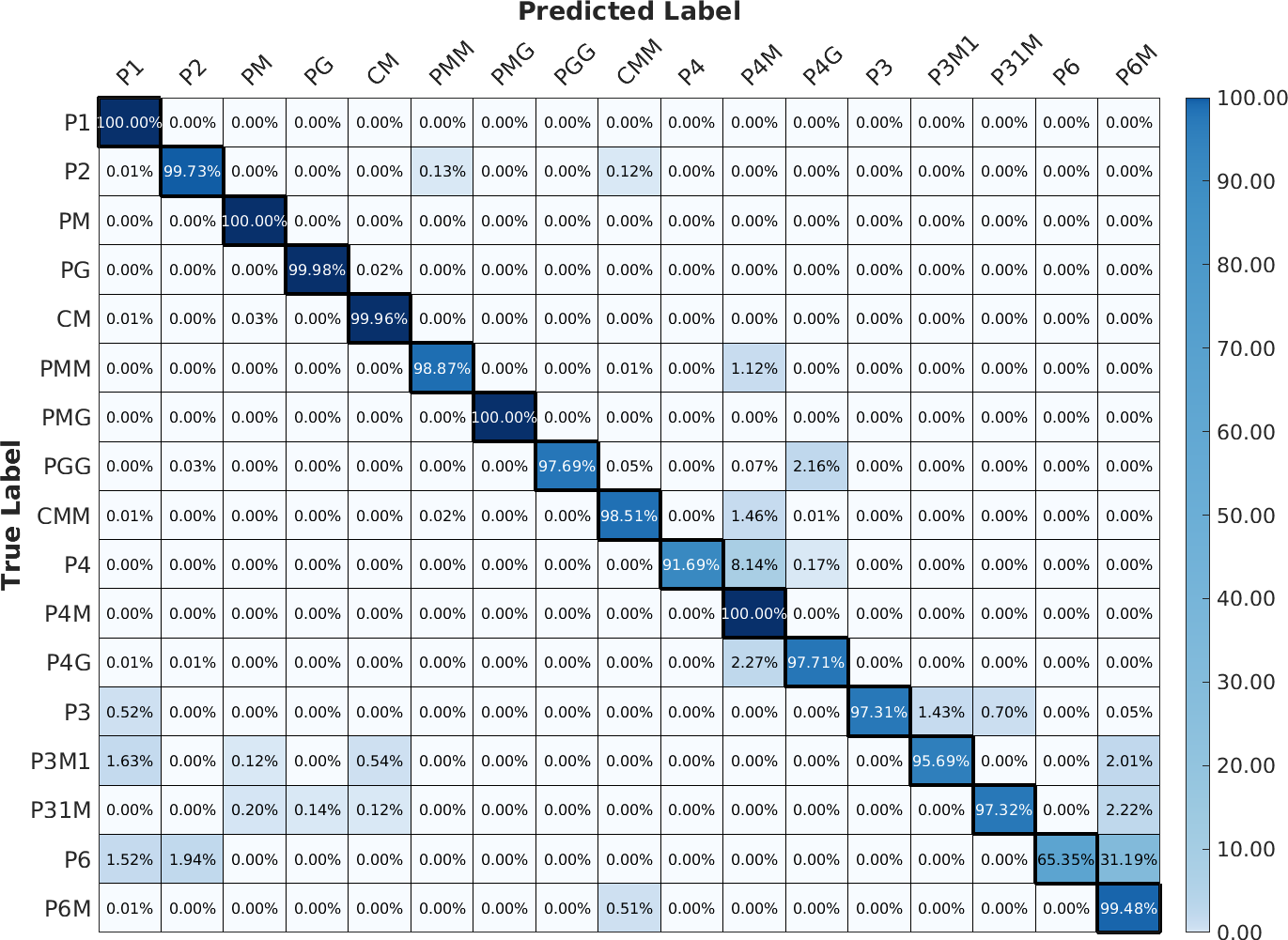} 
	\caption{Confusion Matrix for the deterministic classification algorithm based on the the normalized cross-correlation of the image and a set of transformed images corresponding to the potential symmetries.~\cite{liu2010computational} \label{fig:deterministic CM} }
\end{figure}
}

\subsection{Classifying with Fourier Coefficient}
The Fourier Coefficients transform pixel information to sinusoidal function, redefining the information by the periodicity of the patterns.  If these are easily classifiable then EscherNet might be learning these coefficients for each group rather than the actual symmetries.  

The training data consists of 100 randomly chosen WPIs per group (1,700 total) with all augmentations applied.  The Fourier Coefficients are extracted from each 128x128 pixel WPI and the real and imaginary elements are separated creating 32,768 features per image.  The features' range are normalized between [0,1], ranked using the Augmented Variance Ratio score~\cite{collins2005online,liu2003quantified,liu2003facial,liu2004discriminative}, and the highest scoring 400 features are selected.  These features are then classified 5 times using a multiclass ECOC classifier~\cite{allwein2000reducing,escalera2009separability,escalera2010decoding} with Gaussian SVM learners~\cite{cristianini2000introduction} and tested on the same test dataset as EscherNet.    

The classification obtained a mean classification accuracy of $ 7.1\% \pm 1.1\% $ which is barely above the random chance of 5.9\%.  This shows that the Fourier coefficients features are not sufficiently 
discriminative for the classification of wallpaper patterns. 
(Figure~\ref{fig:avr results 12000}).

\subsection{Comparing Alternate Features with Network}
In order to compare the image features and Fourier coefficient classification with the network, we train a multiclass SVM on the activations from the last layer of EscherNet (before the softmax).  For this experiment, we are using EscherNet as a feature extractor, not a classifier.  This allows us to compare EscherNet's discriminative behavior while also limiting the training dataset 100 WPIs per group and all augmentations applied, the same as the other SVM classifications. 
Using the same classification process as before, these activations are classified 5 times using a multiclass ECOC classifier~\cite{allwein2000reducing,escalera2009separability,escalera2010decoding} with Gaussian SVM learners~\cite{cristianini2000introduction} and tested on the same test dataset as EscherNet.  The mean accuracy over 5 classifications is $97.22\% \pm 3.29\%$, slightly lower than the $98.9\% \pm 1.6\%$ which using the EscherNet classifier with the same testing activations.  These classification accuracies are much higher than either the classifications using image features or Fourier coefficients.  

\subsection{Unsupervised Detection of Wallpaper Patterns}
We evaluate the detections on an unsupervised wallpaper detection algorithm from \cite{liu_etalPAMI2004} we call the \textit{U-Method}.  The algorithm uses the normalized cross-correlation between the image and a transformed version of the image corresponding to the possible symmetries except translation, found via autocorrelation with patches in the image to find the translational shifts.  The possible symmetries are rotation of $60^\circ, 90^\circ, 120^\circ, 180^\circ $, and reflection or glide reflection along the translation shifts of T1 and T2 and on the diagonal of the translational shifts D1 and D2~\cite{liu_etalPAMI2004}.  Each transformation yields a value between [0,1] on the strength of the symmetry in the image.  
Our version of the algorithm uses the distance between the weakest symmetry value which defines each group and the next lowest symmetry which does not define the group, which we call the \textit{margin}.  
This algorithm obtains a classification accuracy of $ 96.4\% \pm 8.3\%$ w
(Figure~\ref{fig:deterministic CM}).  

\subsection{Expanding the Transformations}
\begin{figure*}[ht!]
	\centering
	\resizebox{\linewidth}{!}{ 
	\begin{tabular}{cc}
		All & Adding Reflection \\
		\includegraphics[width=0.3\linewidth]{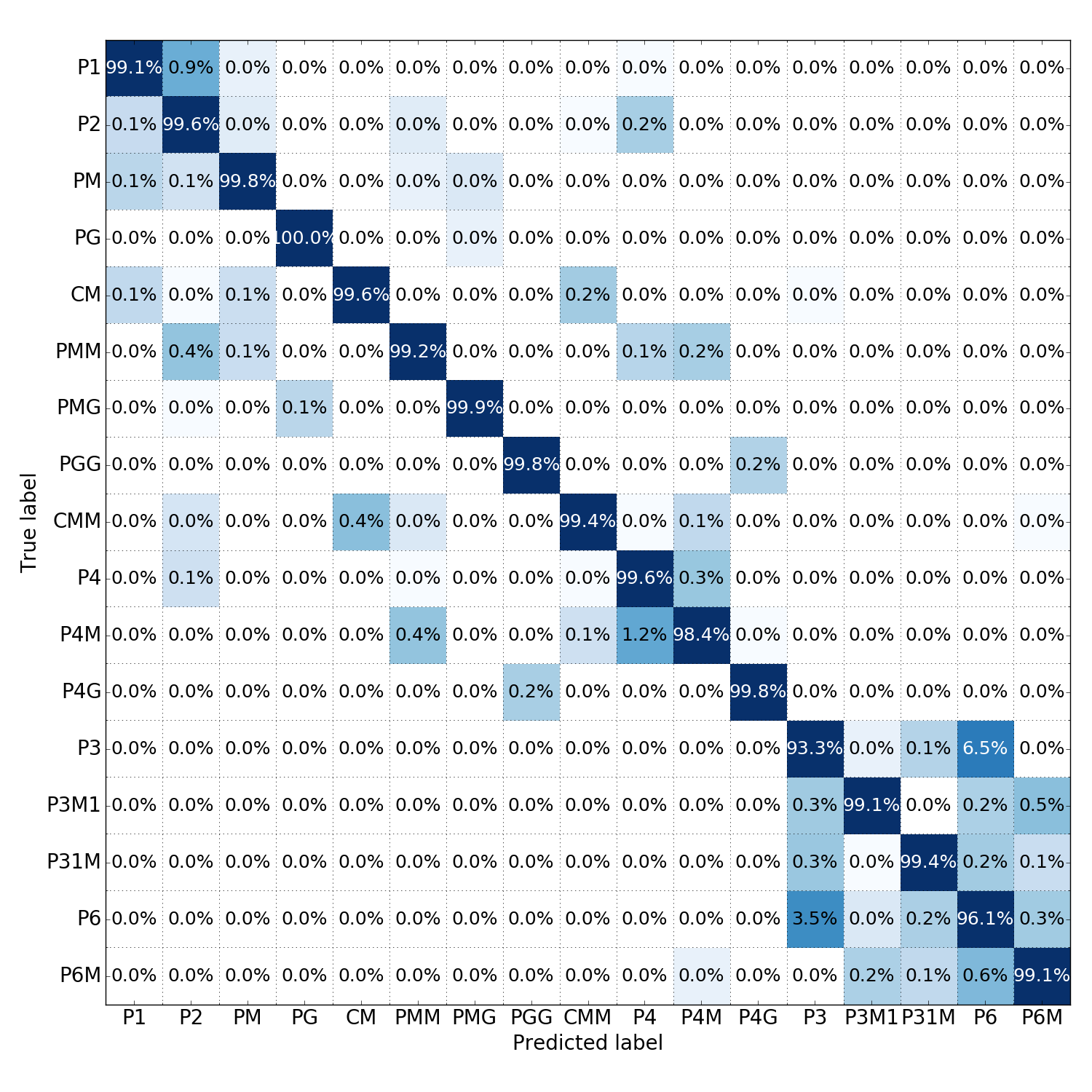} &
		\includegraphics[width=0.3\linewidth]{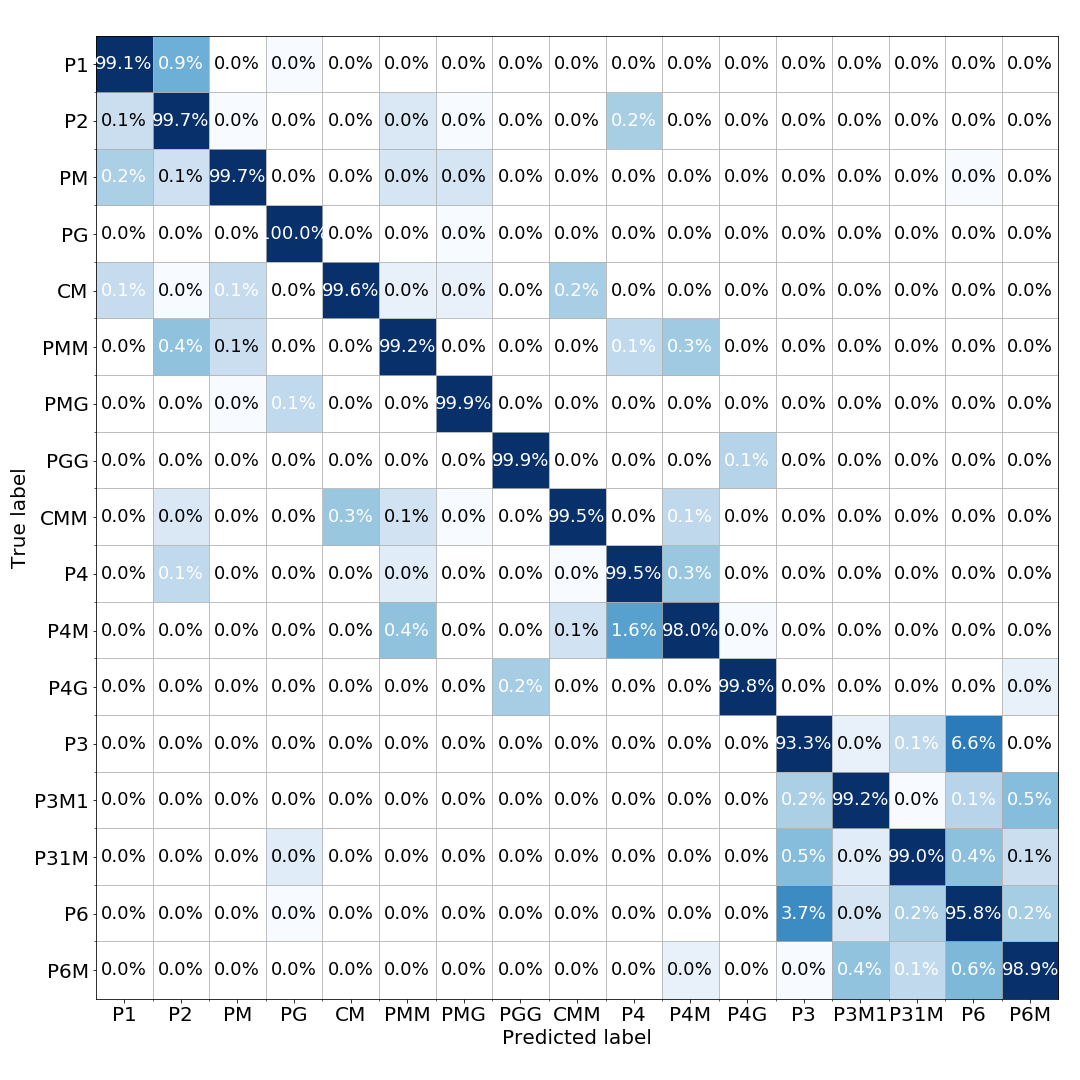} \\
		Expanding Scale Range & \parbox{0.45\linewidth}{\centering Adding Reflection and \\ Expanding Scale Range} \\
		\includegraphics[width=0.3\linewidth]{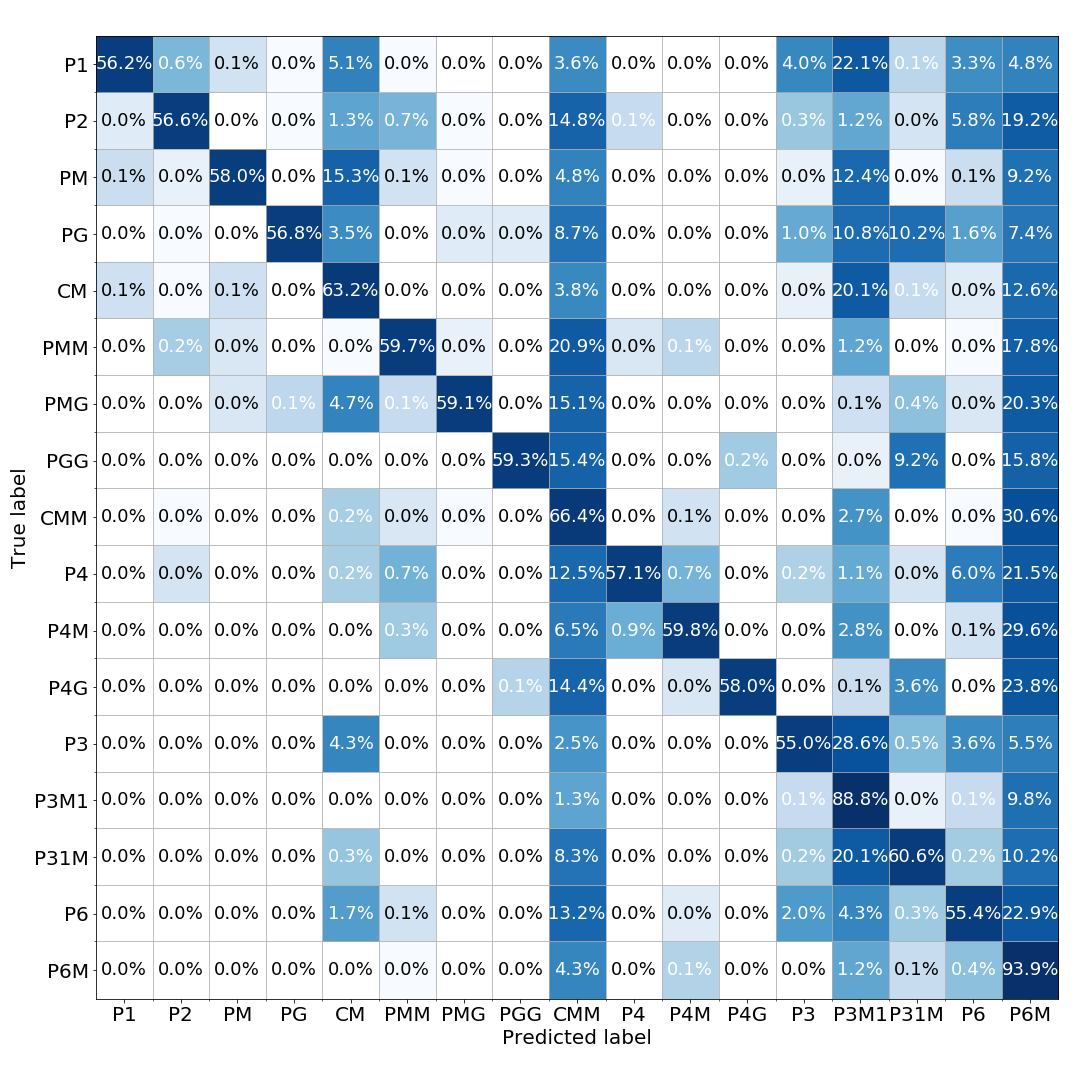} &
		\includegraphics[width=0.3\linewidth]{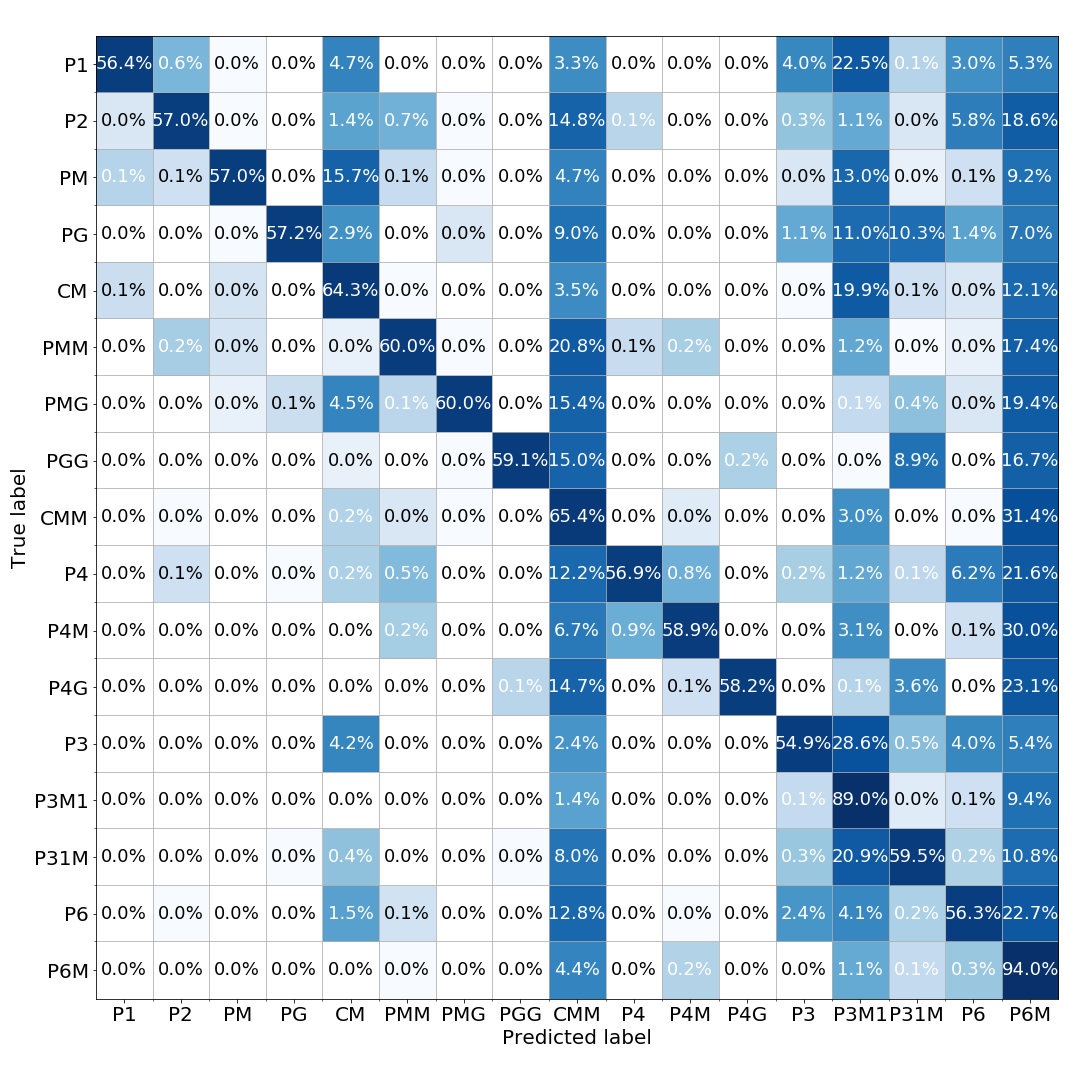}
	\end{tabular}
	}
	\caption{The classification accuracies for the expanded augmentation transformations.   These results show that a reflection augmentation has almost no affect on classification scores are predicted with a increase of 0.1\% (well within the standard deviation).  However, increasing the scale change has a large affect as shown by the dip from 98.7\% to 62.6\% (a dip of 36.1\%).  One oddity is that many patterns are now classified as CMM and P6M (and to a lesser extent CM) regardless of their shared symmetries.  We would need to retrain the networks using allow for this expanded scale. All examples here rotation and translation applied to them in addition a normal scale range if not expanded.  Normal represents all transformation EscherNet is trained on.}
	\label{fig:cm extended transformations}
\end{figure*}

For this experiment we modified the augmentation transformations to extend the range of the scaling and add reflection (both together and separately).  The scale is expanded to having at least an area of $2 \times t_1 \times t_2$ which is $ 64 \times 64 $ pixels from the original image.  We also use random rotation and translation in addition a normal scale range if not expanded for both of these experiments.  We use the original network trained on 48k WPIs per groups.  The classification accuracies are shown in Table~\ref{tab:expanded transformations} and confusion matrices are shown in Figure~\ref{fig:cm extended transformations}.  

These results show that a reflection augmentation has almost no affect on classification scores predicted with a dip of 0.1\% (well within the standard deviation).  However, increasing the scale change has a large affect as shown by the dip from 93.6\% to 59.1\% (a dip of 34.5\%).  One oddity is that many patterns are now classified as CMM and P6M (and to a lesser extent CM) regardless of their shared symmetries.  We would need to train EscherNet to allow for this expanded scale. 
\COMMENT{
\begin{table} \centering
	\caption{Accuracies while varying the augmentations applied to the testing set.  All examples here rotation and translation applied to them in addition a normal scale range if not expanded.  Normal represents all transformation EscherNet is trained on.  This EscherNet is trained on 48k WPIs per group. \label{tab:expanded transformations}}
	\resizebox{\linewidth}{!}{ 
	\begin{tabular}{ccccc} \toprule
		Augmentations & Normal & Reflection & \parbox{2.25cm}{\centering Expanded Scale} & \parbox{2.25cm}{\centering Reflection + Expanded Scale} \\ \midrule
		\parbox{2.4cm}{\centering Classification Accuracy \\$ \pm $ Class STD} & $ 98.7\% \pm 5.2\% $ & $ 98.8\% \pm 1.7\% $ & $ 62.6\% \pm 10.9\% $ & $ 62.6\% \pm 10.9\% $ \\ \bottomrule
	\end{tabular}
	}
\end{table}
}

\subsection{Comparing our wallpaper generator with the original} \label{sec:generator}
While creating the larger dataset with over a million images, 
there are five main differences between our wallpaper pattern generator and that of 
\cite{kohler2016representation}:\\
 (1) We normalized the unit lattice size.  The old generator equalizes a rectangular patch around each unit lattice.  Since the 3-Fold, 6-Fold, CM, and CMM are generated with diamond shaped lattices, this reduces the area of each of these unit lattices to half the size of the other wallpaper groups.  We scale the size of generation of the repeating patch such that the unit lattice is of the same size for all of the groups and then resize it again to try and make the unit lattice areas as close as possible.  (2) We remove the circular mask which was to prevent humans from orienting the patterns from the edges of the pattern.  In our network, we are using the assumption that the pattern is repeating out infinitely in all directions and this is a crop of that pattern.  (3) We reduce the size of the overall pattern from 600x600 to 256x256 and the unit lattices from 100x100 to 32x32 since the input of our network is 128x128 and we would have to reduce the size of the images as a preprocessing step anyway. (4) We reduce the smoothing of the patterns during the filtering from a 9x9 to a 3x3 (proportional to the reduction in the unit lattice). (5) The generator normalizes the amplitudes of the pattern with the median of the Fourier Coefficients for each individual group.  This makes the problem intractable for our larger dataset so instead we use the mean of Fourier Coefficients in batch sizes of 500 images.  

\begin{table*} \centering
	\caption{Accuracies on WPIs from the Original and Adapted wallpaper generator.  These results are on 100 WPIs per group in each of the train and test sets. The adapted generator generator normalizes the wallpaper pattern WPI area is more separable than the original one. The old generator patterns were generated images at 600x600 with the rectangle around each unit lattice having an area of 100x100, a gray circular mask is applied to each image, the generator uses the median between the Fourier Coefficients, and during filtering the sigma was larger.  We test the old generator by changing the image size to 256x256 with the rectangle around the unit lattice being 32x32 (smaller size), normalizing the unit lattice.  The adapted generator (the one for this paper) generated WPIs of 256x256 with all unit lattices having an area of 32x32.   The alterations to the generators are mainly necessitated by the creation of the larger dataset.  The detailed reasons for all changes are listed in Section~\ref{sec:generator}. \label{tab:generators}}
	\resizebox{\linewidth}{!}{
	\begin{tabular}{ccccc} \toprule
		& Old Generator & 
		\parbox{2.25cm}{\centering Smaller Size} & 
		\parbox{2.25cm}{\centering Normalizing Unit Lattice} & 
		\parbox{2.25cm}{\centering Adapted Generator} \\ \midrule
		\parbox{2.4cm}{\centering Classification Accuracy \% \\(mean$ \pm$STD)} & $56.91\pm30.17$ & $61.58\pm30.30$ & $65.05\pm28.59$  &  $72.53\pm24.93$  \\ \bottomrule
	\end{tabular}
	}
\end{table*}

\begin{figure*}
	\centering
	\resizebox{\linewidth}{!}{
	\begin{tabular}{ccc}
		Original Generator & \parbox{2.5cm}{\centering Smaller Size}& \parbox{2.25cm}{\centering Adapted Generator} \\
		\includegraphics[width=0.3\linewidth]{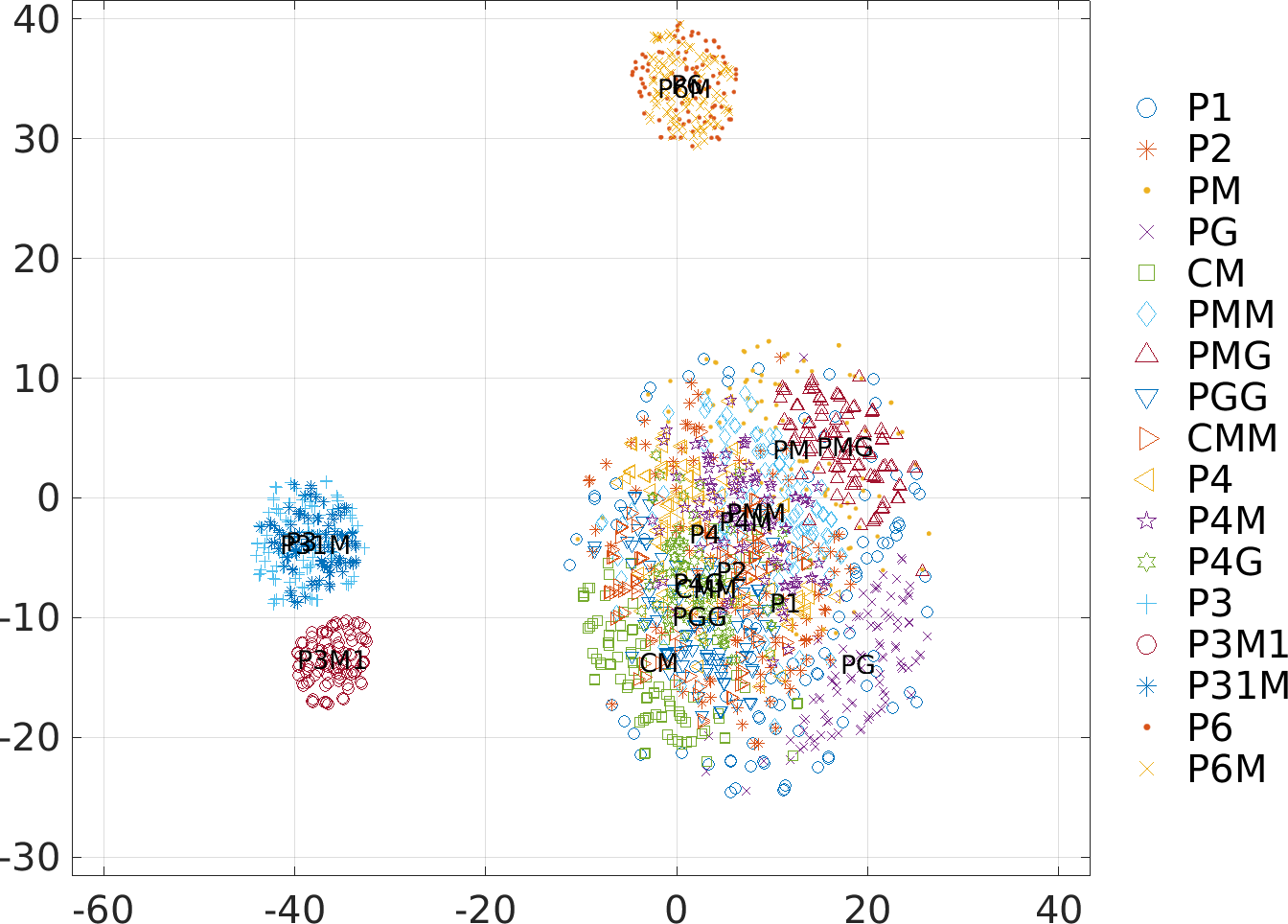} &
		\includegraphics[width=0.3\linewidth]{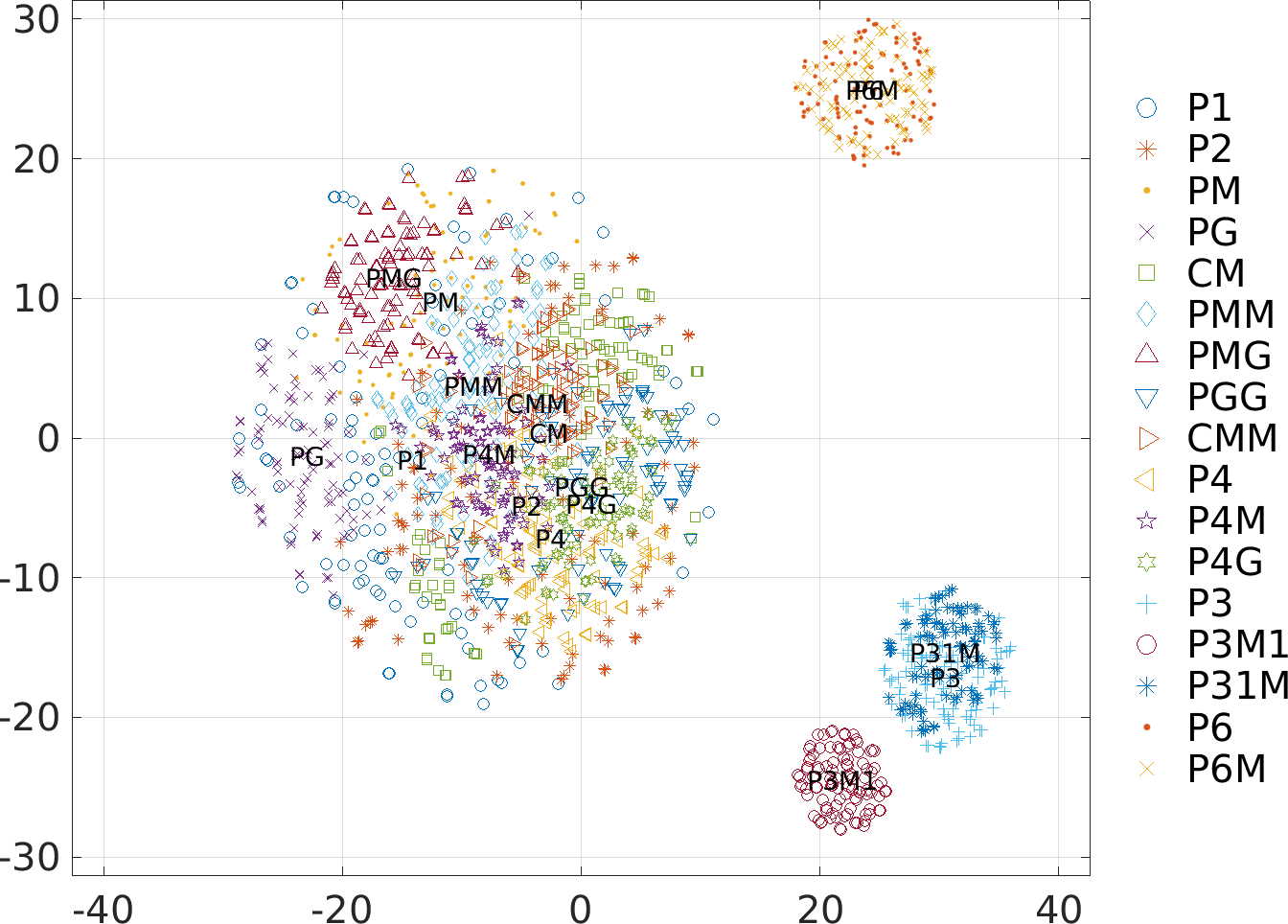} &
		\includegraphics[width=0.3\linewidth]{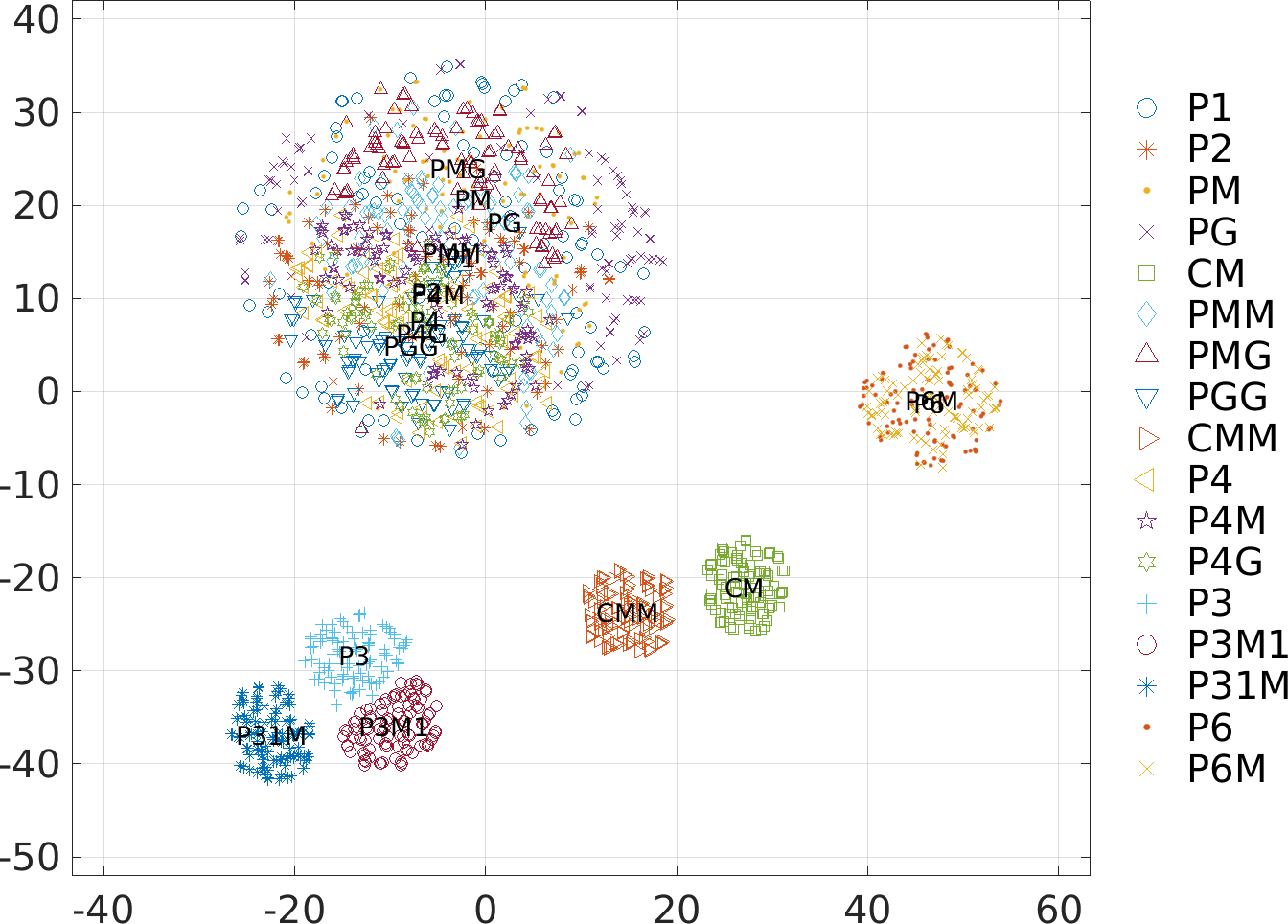} 
	\end{tabular}
    }
	\caption{t-SNE reduction of the different un-augmented datasets.  The adapted generator which normalizes the wallpaper pattern WPI area is more separable than the original one while the old generator results are pretty similar between the two sets of parameters.    The alterations to the generators are mainly necessitated by the creation of the larger dataset.  The detailed reasons for all changes are listed in Section~\ref{sec:generator}.  \label{fig:tsne generators}}
\end{figure*}

\begin{figure*}
	\centering
    \resizebox{\linewidth}{!}{
	\begin{tabular}{ccc}
		Original Generator & \parbox{2.5cm}{\centering Smaller Size}& \parbox{2.25cm}{\centering Adapted Generator} \\
		\includegraphics[width=0.3\linewidth]{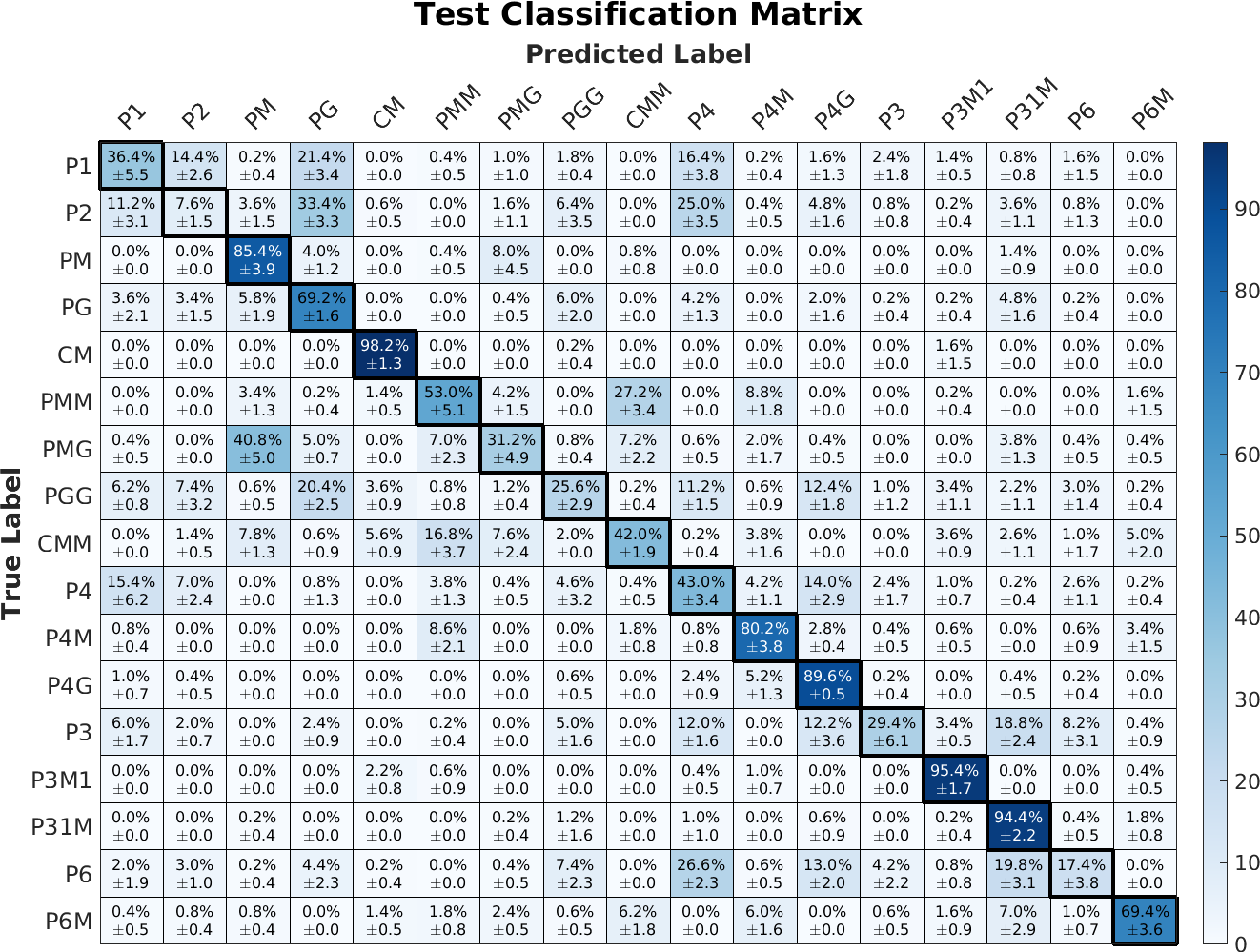} &
		\includegraphics[width=0.3\linewidth]{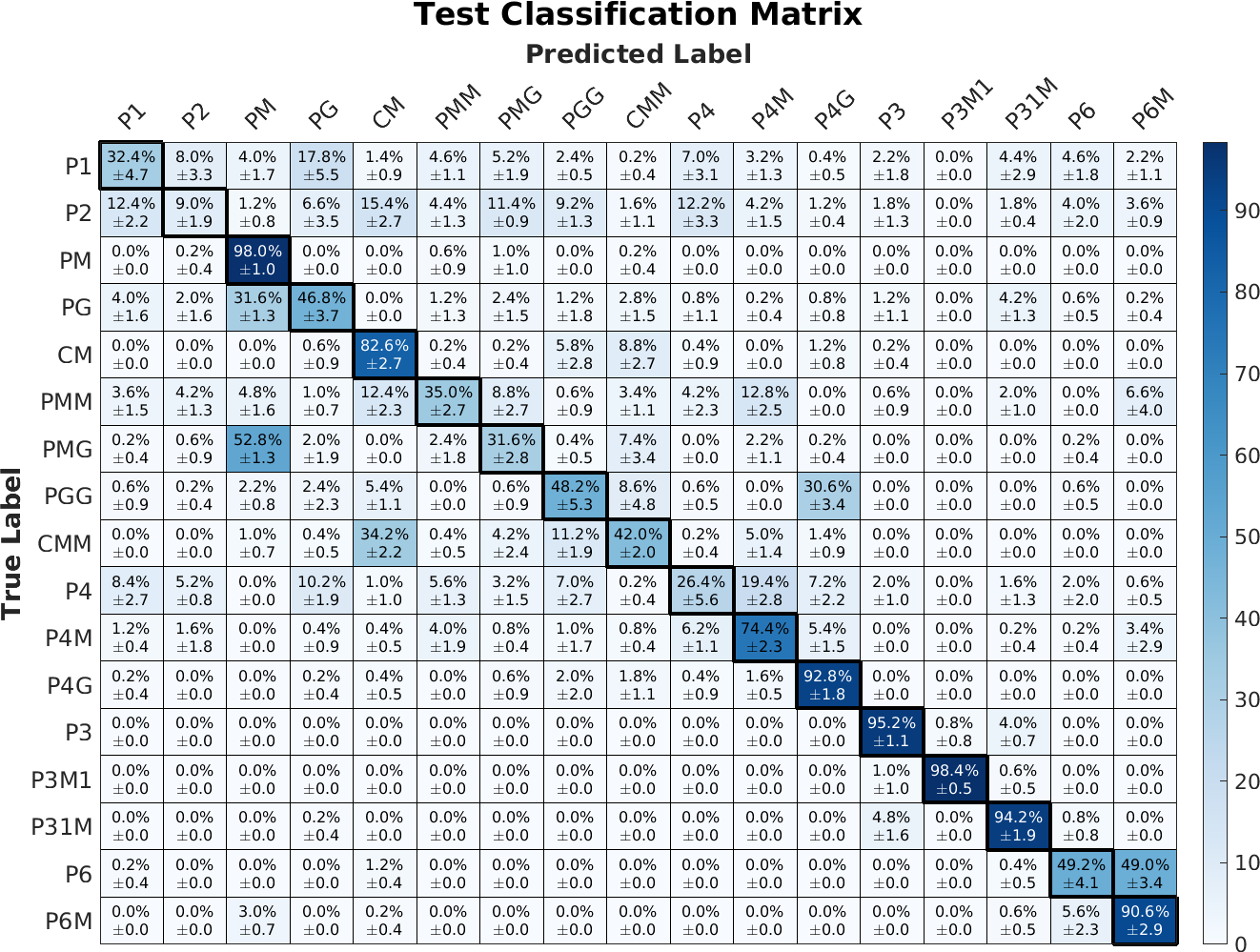} &
		\includegraphics[width=0.3\linewidth]{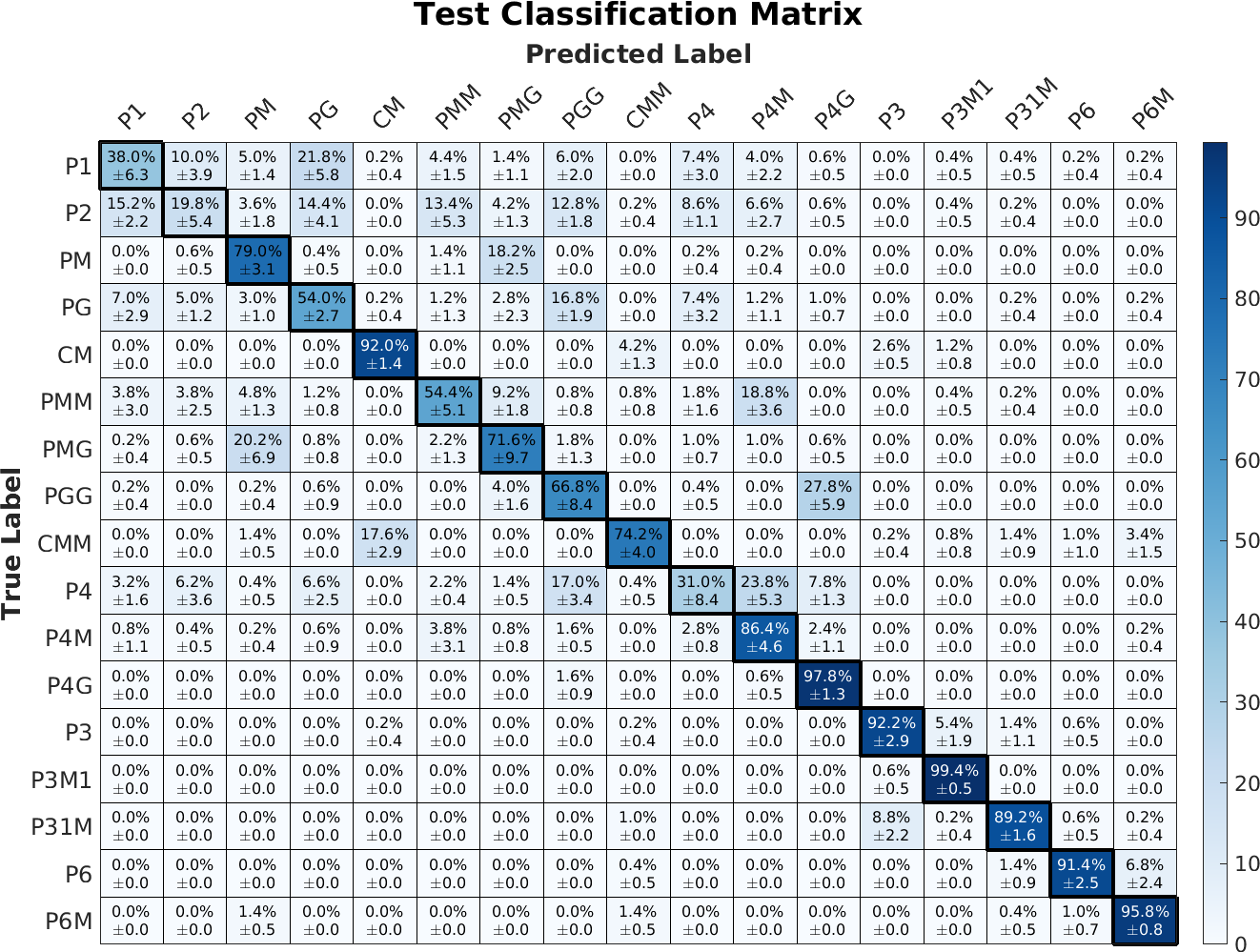} 
	\end{tabular}
    }
	\caption{The classification accuracies from the original and adapted wallpaper generator.  These results are on 100 WPIs per group in each of the train and test sets.  The old original patterns were generated images at 600x600 with the rectangle around each unit lattice having an area of 100x100.  The new generator (the one for this paper) generated WPIs of 256x256 with all unit lattices having an area of 32x32. The alterations to the generators are mainly necessitated by the creation of the larger dataset.  The detailed reasons for all changes are listed in Section~\ref{sec:generator}.}
	\label{fig:generators}
\end{figure*}

In order to compare the original generator and the adapted generator, we first use the BoW Surf for feature and the ECOC multiclass SVM for classification.  We train and test with 100 WPIs per class with no augmentations.  This means that the symmetries between all the images are lined up correctly.  On the original generator's WPIs, we obtain a classification accuracy of $56.91\%\pm30.17$.  When only changing the original generator's image to 256x256 and rectangle around the unit lattice to 100x100, we obtain an accuracy of $61.58\%\pm30.30$.  Next we alter the old generator to have equal sized unit lattices and to create the smaller images and we obtain an accuracy of $65.05\%\pm28.59$.  The confusion matrices and classification scores are shown in Figure~\ref{fig:generators} and all results are shown in Table~\ref{tab:generators}.  Finally, we also test our adapted generator with the same experimental paradigm and obtain a classification rate of $72.53\%\pm24.93$ showing that our data is more separable.  
These accuracies are much higher than the earlier experiment in Section~\ref{sec:Image Features} where we used image features on the augmented data.  This shows the need for augmenting the dataset to help prevent image features from playing a large role in the classification.

We also trained an EscherNet on the 100 WPIs per group for training and testing created with the original generator with no mask, the same size images, and equal sized unit lattices as the new generator to see how EsherNet handles the differences.  We use the same augmentations as the other EscherNets. This EscherNet obtained a classification rate of $57.5\% \pm 20.2\%$.  Future work will delve into why there is a drop in performance here. 

These results show the necessity of augmenting the dataset since the patterns are separable through only image features without them.  Another demonstration of this is shown in Figure~\ref{fig:tsne generators} with a t-SNE reduction of the different datasets.  The new generator which standardizes the different groups is more separable than the old generator.

\begin{figure}[ht]
	\centering
	\includegraphics[width=0.9\linewidth]{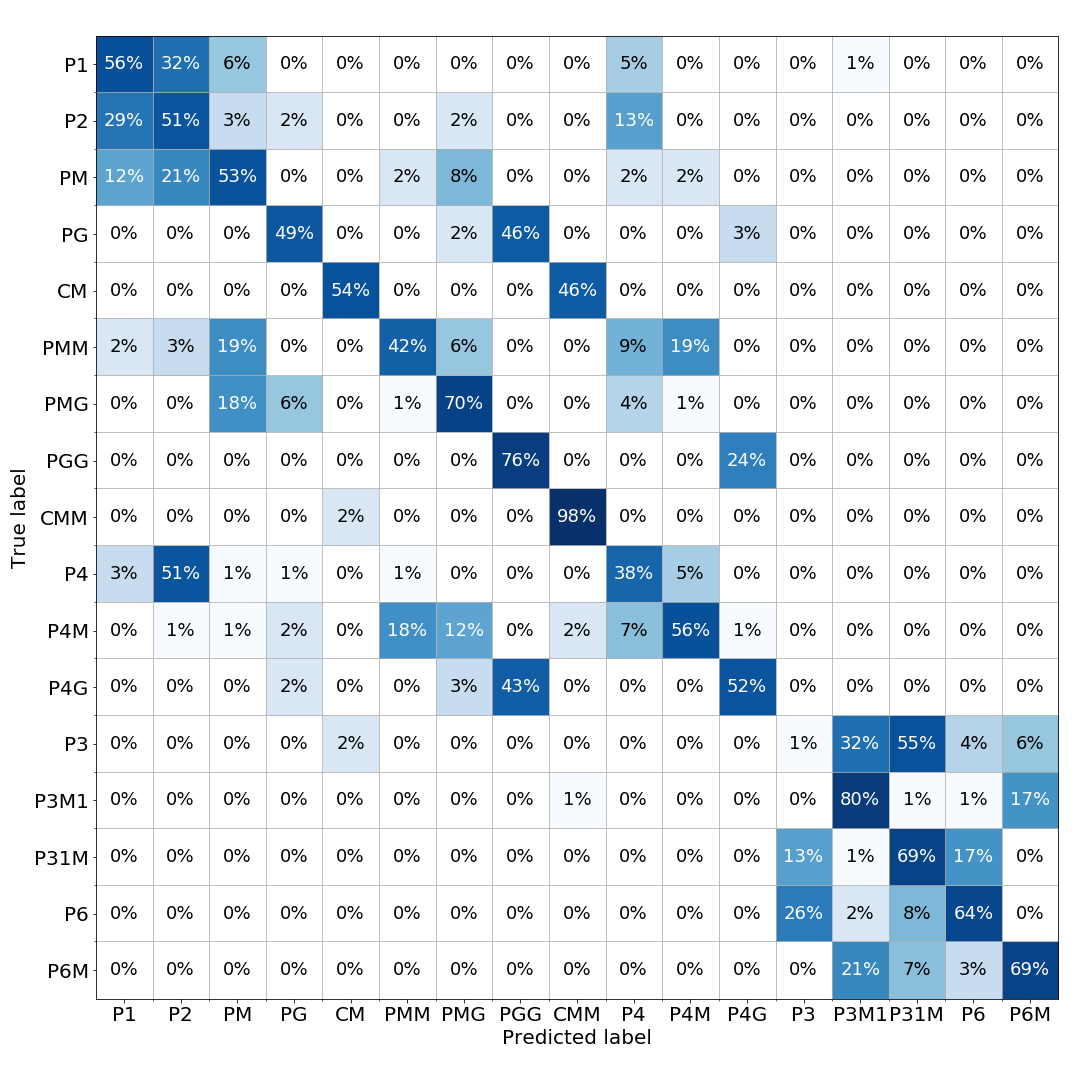}
	\caption{EscherNet trained on 100 WPIs from the original generator with no mask, equal sized lattices, and images of the size 256x256 with a unit lattice with an area of 32x32. }
	\label{fig:EscherNet on original generators}
\end{figure}

\subsection{Adding Patches to P2 Group}
\begin{figure}[ht]
	\centering
	\includegraphics[width=.9\linewidth]{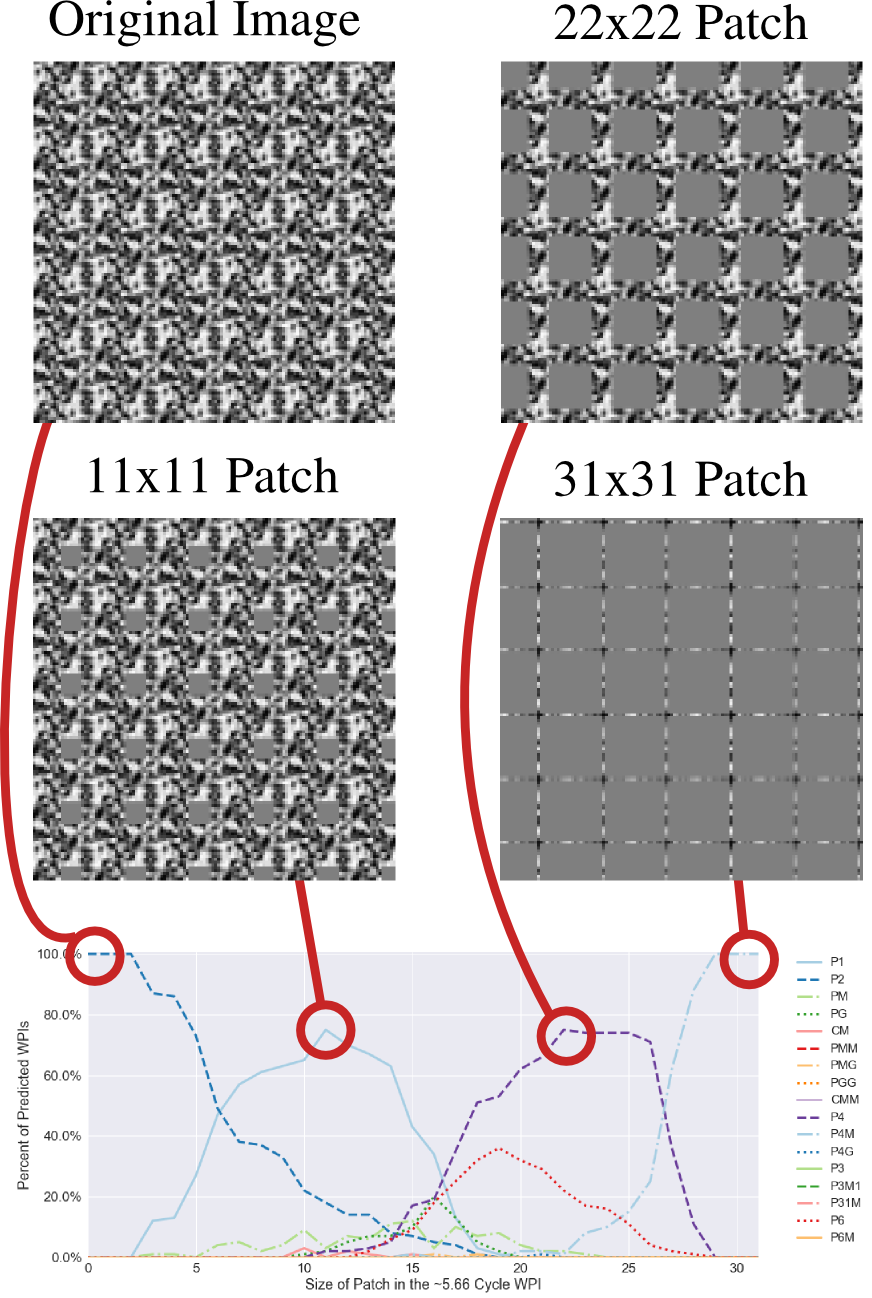}
	\caption{The classification scores while adding a patch to an image. \textbf{Top 2:} The original image and 3 examples of adding different sized patches to the P2 WPIs.   \textbf{Bottom:} How 100 P2 WPIs with mean patchs added to the image are classified.  At first it correctly classifies them as P2 but, even with a small patch size, the patterns become classified as P1.  As the patches starts to become dominant, other labels such as P4 and eventually P4M which might be due to the fact that the Patches are themselves 4-Fold symmetric.  \label{fig:patches}} \vspace{10pt}
\end{figure}

In order to see if the network is actually learning symmetry, we perform an experiment where we add a patch of the mean pixel values at the same place in each unit lattice of a P2 Pattern (only containing 2-Fold rotation symmetry) to try and break the symmetry of the pattern.  All images were correctly identified as the pattern before the 
patch was added.  The center of the patch is at least 10 pixels away from any rotation symmetry center to prevent the patch from being rotationally symmetric. The patch is gradually enlarged until it covers most of the image (Figure~\ref{fig:patches}).   
The patches are inserted into the original image and then pre-processed to have $ \approx5.66 $ cycles (or unit lattices in) in the WPI (the largest scale possible which can allow for all rotations). 

We classify these images using the network trained on 48k WPIs to see how this progressive removal of the 2-fold rotation symmetry affects the classification.  At first it correctly classifies them as P2 but, even with a small patch size, the patterns become classified as P1.  Though as the patches starts to become dominant, other labels such as P4 and eventually P4M might be due to the fact that the Patches are themselves 4-Fold symmetric.  

\subsection{Scale effects}
\begin{figure*}[ht!]
	\centering
	\includegraphics[width=1\linewidth]{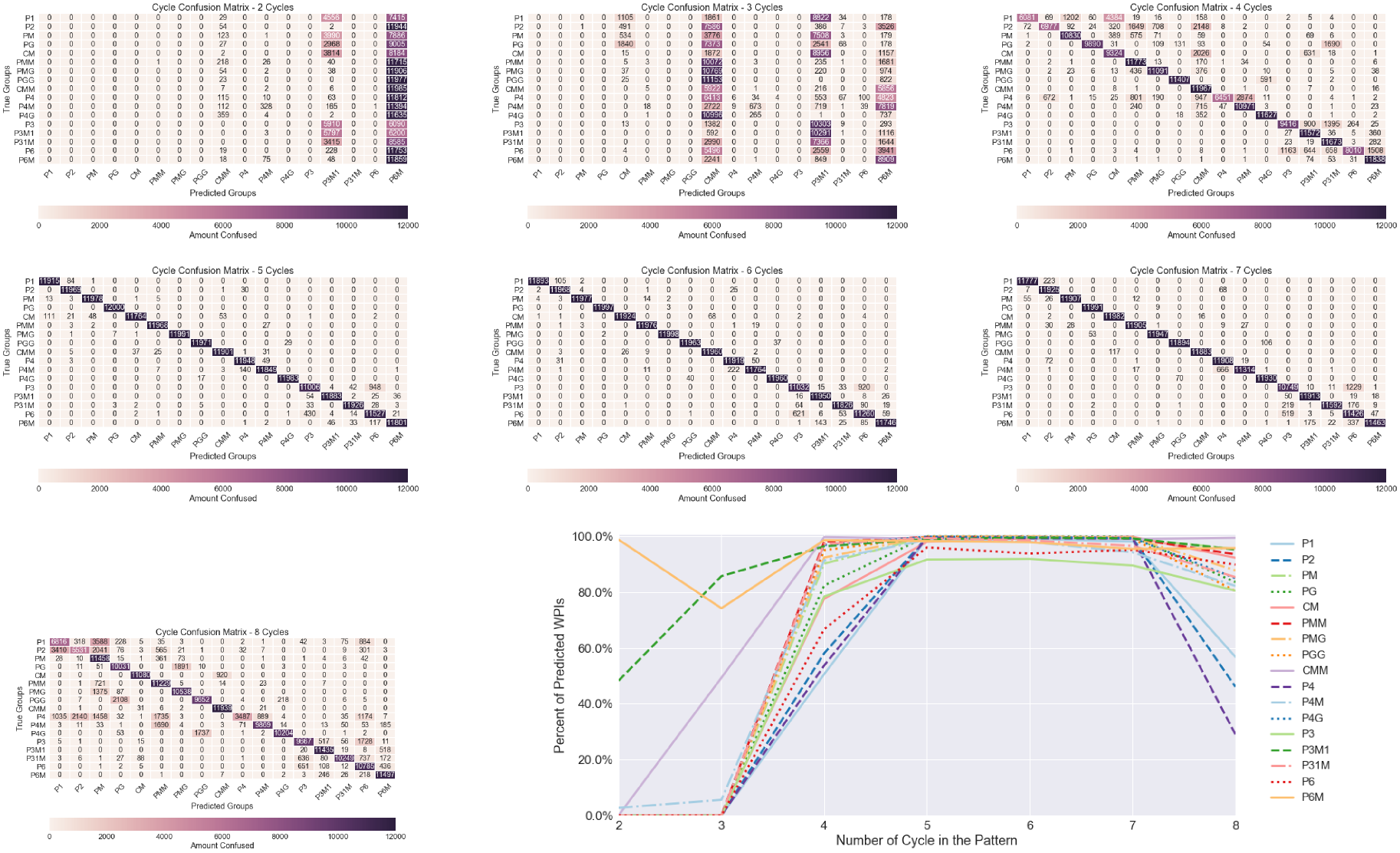}
	\caption{The confusion matrices for different scales representing 2-cycles (400\% scale showing 2x2 unit lattices) to 8-cycles (100\% scale of the original 256x256 WPI).  We use the EscherNet that has been trained on 48k WPIs and with training augmentations of 4-cycles to 8-cycles (100\% to 200\% scaling).  Predictably, EscherNet's accuracy becomes worse at the edges of the training scaling range and does not work outside of the training scale range.  \label{fig:cycles}}
\end{figure*}

We look at how changing the scale in the image affects classification in Figure~\ref{fig:cycles}.  The images are scale to have an area of 2x2 unit latices (64x64 pixels from the 256x256 WPI) to having the area of 8x8 unit lattice (the entire 256x256 WPI).  We use the EscherNet that has been trained on 48k WPIs and with training augmentations of 4-cycles to 8-cycles (100\% to 200\% scaling).  Predictably, EscherNet's accuracy becomes worse at the edges of the training scaling range and does not work outside of the training scale range.

\section{Micro-level: Visualization of Learned Filters}

\noindent{\bf Class Activation Maps}: \\
To understand at a micro-level what EscherNET has learned from the wallpaper patterns, we apply the Class Activation Maps (CAM)~\cite{zhou_etal_CVPR2016learning} used in the Grad-CAM method~\cite{selvaraju2016grad} 
to visualize what parts of the image affect the classifications the most (Figure~\ref{fig:data}).  
Similar to \citet{zeilerECCV2014visualizing}, CAM finds the most influential areas of the image for classifying an image into a particular group by the gradient of the final layer's class activation.  
The discrete and even distribution of the peaks in the input images justifies that the EscherNET is learning translation symmetry generators, since the smallest distances correspond to the translation vector-lengths. The fact that the peaks are
not always aligned with the rotation centers also indicates that translation subgroups are not localized.

\begin{figure*}[t!]
	\centering
	\newlength{\maxImgSize}
	\setlength{\maxImgSize}{.17\linewidth}
	\includegraphics[width=1\linewidth]{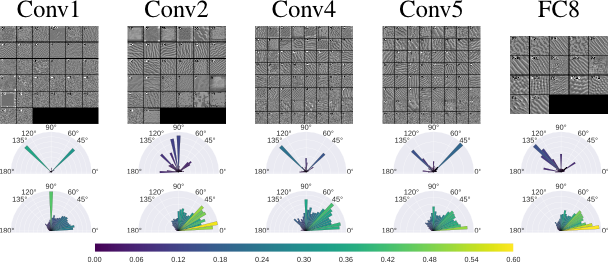}
	\caption{
		TOP: Images created by maximizing the activations for filters at each layer of the EscherNet, and sorted by their scales. 
		At FC8, each wallpaper has a corresponding filter. 
		MIDDLE:  the dominating orientation of the filters are shown. 
		BOTTOM: the histogram of minimum filter angle differences.  
		Observation: from lower to higher layers, the filters evolve from single direction oriented ones (oriented filters) to 
		more complex multi-oriented and multi-scaled ones. it is especially evident at the FC8, where all filters correspond to wallpaper groups P4, P4g and P4M have 4-fold (90 degrees) rotation symmetries.
	}
	\label{fig:max_activations}
\end{figure*}

\noindent{\bf Maximizing Neural Activations}:\\
To further understand what 
each neuron at each layer is ``looking for'' (Figure~\ref{fig:max_activations})~\cite{simonyan2013deep, mahendran_vedaldi_IJCV2016_visualizing,yosinski2015understanding}, 
we maximize each neuron's activation by 
(1) starting from a random input image patch, back propagating the activation and adding it into the image repeatedly for each neuron with an $ \ell_2 $ loss function for 500 iterations;  
(2) using the same parameters as \citet{mahendran_vedaldi_IJCV2016_visualizing} but avoiding 
regularizing the image since the common approach of using LP and TV regularization~\cite{simonyan2013deep,mahendran_vedaldi_IJCV2016_visualizing} destroys the patterns in the synthesized images created with the convolution layer filters;
(3) using a Hough Transform~\cite{duda1972use} to find the orientation of the gradients
in each filter.   
We compute a Softmax normalized histogram of the orientations at each filter to discover the most dominant angles of orientation;
(4) estimating the scale of filter patterns using autocorrelation and the distance between the closest peaks
to determine the scale; 
(5) finally, sorting the filters by their scales for all layers except for the FC8 layer where each filter corresponds to 
a specific wallpaper group. 

From Figure \ref{fig:max_activations}, we have observed that from lower to higher layers the filters evolve from single direction oriented ones (oriented filters) to 
more complex multi-oriented and multi-scaled ones. This is especially evident at FC8, where all filters corresponding to wallpaper groups P4, P4G and P4M have 4-fold (90 degrees) rotation symmetries.

%%%%%%%%%%%%%%%%%%%%%%%%%%%%%

\section{Discussion and Future Work}
\label{Discussion}

{\bf Learning subgroup hierarchy}: 
As a shallow, generic artificial neural net,  
EscherNet performs surprisingly well. Not only does it learn to discriminate the 17 wallpaper groups (Table \ref{tab:aug classification matrix}), it also learns a significant portion of the nontrivial topological structure corresponding to the mathematically proven WGH (Figure \ref{fig:WP_Hierarchy}).
Even though the input images are relatively clean (synthesized from a random noise patch), they also provide a well-controlled computational 
environment to ask basic questions such as whether symmetry groups can be learned directly from the data in contrast to the classic first principles approaches in computer vision where rules are built into the algorithm already. The affirmative response from EscherNet 101 (a baseline) is encouraging since it suggests that it 
might be feasible to use a machine-learning based neural-net framework as a crude model for pattern/texture perception, in particular for 
regular or near-regular textures. 
Echoing one of the findings from \cite{kohler2016representation}, rotation symmetries 
seem to be one of the confusers at the lower layers (lower level) of EscherNet (e.g. P3 and P6 classification rates are the lowest, and mutually confused).

{\bf Asymmetrical Similarity}: 
The confusion matrix of EscherNet (Figure \ref{fig:confusion matrix}) is particularly interesting to examine because the off-diagonal places where errors are made the most happen to be between pairs of groups that have a subgroup relationship. 
Furthermore, the errors made are {\em asymmetrical} and much higher in the direction when classifying from a more general group (subgroup) to a more specific group (a super group). This is precisely the key issue in hierarchical class relations (not a flat, mutual exclusive set of classes), and often handled incorrectly. These types of asymmetrical
similarities have been discussed by researchers previously in human 
perception (\citet{tversky1977features}) and computer vision (\cite{kanatani1997comments} among others). 
The observation that EscherNet has erroneous output where mistakes are expected is worth investigating.

{\bf Future work}: We plan to extend this work in at least two directions. One is to study the wallpaper group transition under affine deformation since our earlier work \cite{liu2001skewed} suggests that the 17 wallpaper groups migrate into clusters following provable rules. If an EscherNet 102 version can learn such rules under affine deformation, we expect fruitful applications in computer vision using an affine camera model. Another direction is to explore different network structures and visualization tools to thoroughly understand when and how the fundamental symmetries and their combinations are learned. Yet another extension is to increase the noise in the data, to verify the limit of learning symmetry groups from real world wallpaper patterns.

\section{Acknowledgement}

This project was funded by an National Science Foundation (NSF) 
{\em Creative Research Award for Transformative Interdisciplinary Ventures} (CREATIV) program, a.k.a. {\bf INSPIRE}: Symmetry Group-based Regularity Perception in Human and Computer Vision (NSF IIS-1248076), and is supported in part by NSF Award IIS-1909315.

	\bibliography{2023_EscherNET_arXiv}
	\bibliographystyle{icml2018}

\vfill
\pagebreak

\end{document}